\def\arxivbuild{}
\documentclass{article}
\usepackage{iclr2027_conference,times}

\usepackage[T1]{fontenc}
\usepackage[utf8]{inputenc}
\usepackage{microtype}
\usepackage{amsmath, amssymb}

\usepackage{graphicx}
\graphicspath{{figures/img/}}      %
\usepackage{subcaption}
\usepackage{booktabs}
\usepackage{tabularx}
\usepackage{float}               %
\usepackage{wrapfig}             %
\usepackage[normalem]{ulem}      %
\usepackage{xcolor}
\usepackage{colortbl}            %
\usepackage{enumitem}
\setlist{itemsep=2pt, topsep=3pt, parsep=0pt, leftmargin=1.4em}

\usepackage{hyperref}            %
\usepackage{url}
\usepackage{titletoc}
\makeatletter
\newcommand{\beginsupplementary}{%
  \clearpage
  \appendix
  \noindent{\LARGE\scshape Supplementary Material\par}%
  \vspace{0.25in}%
  \section*{Contents}%
  \startcontents[supp]%
  \printcontents[supp]{}{1}{\setcounter{tocdepth}{2}}%
  \let\supp@section\section
  \let\supp@subsection\subsection
  \renewcommand{\section}{\clearpage\supp@section}%
  \renewcommand{\subsection}{\if@nobreak\else\clearpage\fi\supp@subsection}%
}
\makeatother

\newif\ifarxiv
\ifdefined\arxivbuild\arxivtrue\fi
\newcommand{\projectpage}{https://segue-video.github.io/}   %
\newcommand{\coderepo}{https://github.com/emirks/SEGUE}         %
\newcommand{\modelpage}{https://huggingface.co/emirkisa/SEGUE}  %
\definecolor{linkblue}{RGB}{58,108,165}
\newcommand{\linkicon}[1]{\raisebox{-0.1em}{\includegraphics[height=0.85em]{icons/#1}}}
\newcommand{\linkitem}[3]{\linkicon{#1}\hspace{0.35em}\href{#2}{\textcolor{linkblue}{#3}}}
\newcommand{\seguelinks}{{\normalfont\hypersetup{pdfborder={0 0 0}}%
  \linkitem{house}{\projectpage}{Project Page}\qquad
  \linkitem{github}{\coderepo}{Code}\qquad
  \linkitem{huggingface}{\modelpage}{Model}}}   %

\newif\ifdraft
\drafttrue

\newcommand{\segue}{SEGUE}                      %
\newcommand{\ctt}{TEG}                          %
\newcommand{\cttlong}{One-Shot Transition Effect Generation}

\newcommand{\refvfx}{refVFX}
\newcommand{\vap}{Video-As-Prompt}
\newcommand{\vfxmaster}{VFXMaster}
\newcommand{\ltx}{LTX-2}

\newcommand{\tsep}{\kern1.5pt\textperiodcentered\kern1.5pt}         %

\colorlet{phc}{blue!65!black}

\newif\ifrevmarks \revmarkstrue
\ifarxiv\revmarksfalse\fi   %
\definecolor{revblue}{RGB}{0,70,200}
\newcommand{\rev}[1]{\ifrevmarks{\color{revblue}#1}\else#1\fi}
\makeatletter
\newcommand{\revfloat}{\ifrevmarks\color{revblue}%
  \def\fnum@table{{\color{revblue}\tablename~\thetable}}%
  \let\revfloat@caption\caption\renewcommand{\caption}[1]{\revfloat@caption{{\color{revblue}##1}}}\fi}  %
\makeatother
\newcommand{\tabsec}[2]{\rowcolor{black!6}\multicolumn{#1}{l}{\emph{#2}}}  %

\newcommand{\lerp}{\ell}                                    %

\makeatletter
\renewcommand{\paragraph}{\@startsection{paragraph}{4}{\z@}{0.375ex plus 0.125ex minus .05ex}{-1em}{\normalsize\bf}}
\makeatother

\ifarxiv\iclrfinalcopy\fi

\title{\ifarxiv\centering\else\raggedright\fi SEGUE: Visual \underline{S}ignal as \underline{E}xclusive
       \underline{G}\underline{u}idance for One-Shot Transition \underline{E}ffect Generation}

\author{\makebox[\dimexpr\textwidth-2\tabcolsep\relax][c]{\begin{tabular}[t]{@{}c@{}}
Emir Kisa$^{1}$\quad Victor Escorcia$^{2}$\quad Junho Kim$^{1}$\quad Alexander Schwing$^{1}$\\
Ozgur Kara$^{1\dagger}$\quad James Matthew Rehg$^{1\dagger}$\\
\normalfont $^{1}$University of Illinois Urbana-Champaign\qquad $^{2}$MacondoLabs\\[9pt]
\seguelinks
\end{tabular}}}

\begin{document}
\maketitle
\ificlrfinal\begingroup\renewcommand{\thefootnote}{}\footnotetext{$^{\dagger}$Joint last authors.}\endgroup\fi
\ifarxiv\lhead{}\renewcommand{\headrulewidth}{0pt}\fi   %

\vspace{-9mm}
\begin{figure}[H]
  \centering
  \includegraphics[width=\linewidth]{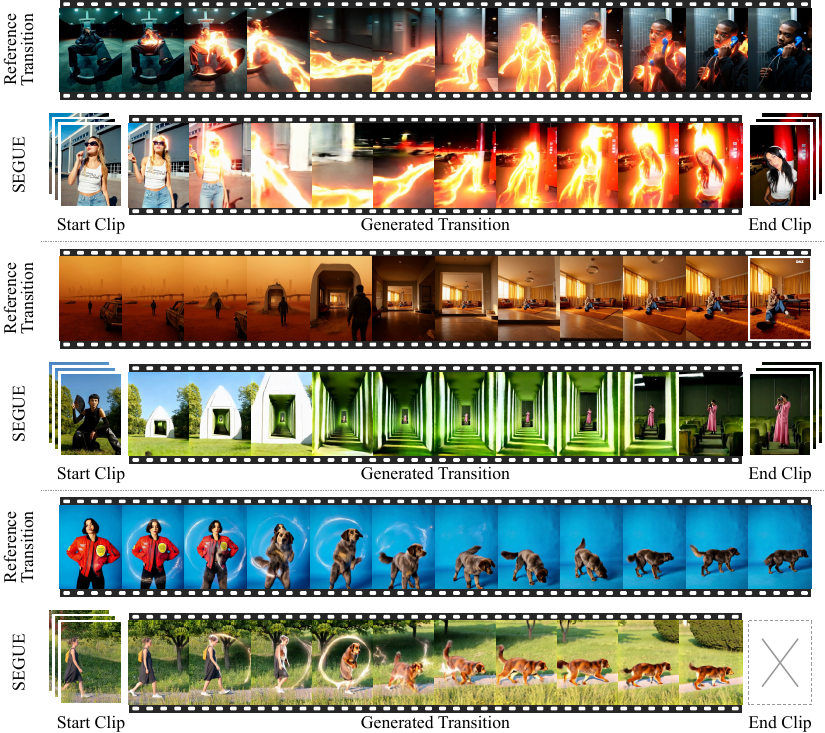}
  \caption{\textbf{\segue{}} is a one-shot transition effect generator: it successfully connects two endpoints (a video clip, a static image, or absent) given a reference transition. \ifarxiv Videos on the project page.\else Videos available in supplemental.\fi}
  \label{fig:hero}
\end{figure}
\vspace{-4mm}   %

\begin{abstract}
One-Shot Transition Effect Generation (TEG) aims to bridge two video clips using a transition effect provided in a reference clip. However, existing methods learn a text shortcut: they rely heavily on prompts with detailed effect descriptions, and the visual transition signal loses its weight. We present SEGUE, a framework designed to read the transition strictly from the visual demonstration via two key mechanisms:
i) \textbf{Visual reference isolation} during training, pairing inputs with effect-free neutral prompts and applying a one-way attention mask that keeps the reference independent of the target; and
ii) \textbf{Null-reference guidance} at inference, which defines the unconditional path as a dissolve of the reference endpoints—adding no motion or material of its own—to isolate and amplify the visual transition effect.
SEGUE flexibly supports endpoints that are video clips, static images, or absent, unifying two-sided TEG and single-sided visual effect transfer within a single model. We also introduce the SEGUE-Benchmark, a cross-paired dataset of 1,692 transition classes and 56,368 training pairs spanning creative presets, parametric procedural transitions, and visual effects. On held-out effects, SEGUE outperforms baselines adapted to TEG and previous visual effect transfer methods even without a detailed effect description in the prompt, serving as the first open-source framework for one-shot TEG.   %
\end{abstract}

\section{Introduction}
\label{sec:intro}

Transitions shape the pacing and continuity of a video, deciding how one shot gives way to the next~\citep{zhang1993automatic,soucek2024transnet,shen2022autotransition}. Beyond simple cuts or dissolves, a \emph{transition effect} performs this hand-off with its own visual dynamics, such as volumetric smoke, opening portals, or wipes, and brings in motion and material that belong to neither endpoint~\citep{liu2025vfx,mao2026omni}. While video diffusion models can bridge clips~\citep{chen2023seine,yang2025versatile,kan2026sage,hacohen2026ltx}, transferring an arbitrary transition effect from a reference clip remains challenging. The model must connect unrelated scenes and transfer the effect's dynamics without leaking the reference's own scenes into the output~\citep{li2025vfxmaster,jones2026tuning,yin2026vcbench}. We formalize this task as \cttlong{} (\ctt{}, Figure~\ref{fig:hero})\footnote{A transition effect goes from the start clip to the end clip, whereas a visual effect is conditioned on the start endpoint alone and ends freely, as in the last row of Figure~\ref{fig:hero}. We call the first task \ctt{} and the second visual effect transfer.}: given a reference clip showing a transition effect and a target pair of start and end clips, synthesize that effect between them.

\begin{figure}[t]
  \centering
  \includegraphics[width=\linewidth]{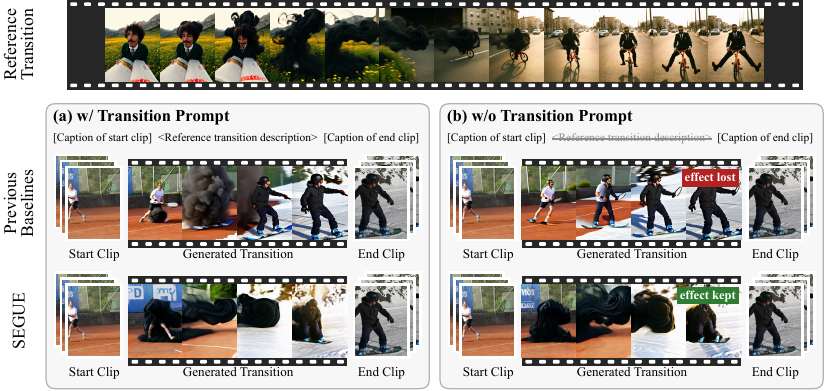}
  \caption{\textbf{The text shortcut.} Given the same visual reference (top) and endpoints, previous methods rely heavily on the effect prompt and under-utilize the visual reference; with the neutral prompt (b), they fail to produce the transition and degrade in quality. In contrast, \segue{} reads the dynamics directly from the visual reference, preserving the effect with the effect prompt (a) and with the neutral prompt (b).}
  \label{fig:motivation}
\end{figure}

Current reference-conditioned approaches are primarily developed for visual effect transfer onto a single starting frame or a full video~\citep{li2025vfxmaster,bian2026video,jones2026tuning,yang2026effectmaker}. They pair the visual reference with a text prompt that describes the effect in detail, which we call the effect prompt. This creates a \textbf{text shortcut}: because text provides a compact semantic signal, models learn to infer the effect from the prompt, leaving the visual reference under-utilized (Figure~\ref{fig:motivation}). As a result, once the effect description is removed from the prompt, the visual reference signal cannot guide the generation and the transition quality drops. More importantly, text cannot convey the exact motion and timing of an effect, and has no words for a novel, uncataloged effect. Because the true dynamics exist only in the video frames, the model must read and transfer the transition directly from the visual demonstration.

To address this limitation, we introduce \segue{}, a framework that reads the transition strictly from the visual demonstration, through visual reference isolation during training and null-reference guidance at inference. During training, \segue{} enforces reference isolation through two complementary designs: neutral text conditioning that contains no description of the intended effect, to close the text shortcut, and a one-way attention mask that allows the target to read the reference while preventing reference representations from being rewritten. At inference time, \segue{} amplifies this visual signal using \emph{null-reference guidance} (NRG). Rather than contrasting against an empty condition, which removes both the effect and the reference scenes and so amplifies both, NRG contrasts against a null reference that keeps the scenes of the reference and removes only the transition effect, built as a dissolve of the reference's own endpoints. Because this dissolve adds no motion or material of its own, steering away from it amplifies the transition dynamics and cancels out the leakage of the reference scenes. Finally, \segue{} supports endpoints that are video clips, static images, or absent, unifying two-sided \ctt{} and single-sided visual effect transfer within a single model.

Because no existing benchmark supports training or evaluating reference-conditioned transitions, we introduce the \textbf{\segue{}-Benchmark}. It establishes a comprehensive dataset spanning creative preset transitions, parametric procedural transitions, such as geometric wipes, volumetric zooms, and fluid warps, and complex physical visual effects. On held-out effects, \segue{} outperforms baselines adapted to \ctt{} and surpasses previous visual effect transfer methods, even though both rely on text prompts containing the effect description, while \segue{} operates on the isolated visual reference alone.

In summary, our key contributions are:
\begin{itemize}
    \item We formalize \cttlong{} (\ctt{}) and introduce \segue{}, the first model designed for one-shot transition effect generation, capable of flexibly transferring transitions across arbitrary endpoints, whether each is a video clip, a static image, or absent.
    \item We propose a two-part method to isolate and amplify the visual reference signal: training-side \emph{visual reference isolation}, which combines a neutral text prompt with a one-way attention mask to close the text shortcut and isolate the visual reference, and inference-side \emph{null-reference guidance}, which contrasts against a dissolve of the reference endpoints to amplify the transition dynamics while suppressing the leakage of the reference scenes.
    \item We build the \segue{}-Benchmark: a cross-paired dataset of 1,692 transition classes and 56,368 training pairs, spanning creative presets, parametric procedural transitions, and visual effects, together with an evaluation dataset and the \emph{transition fidelity} metric.
\end{itemize}
\section{Related Work}
\label{sec:related}

\paragraph{Controllable video generation.}
Pretrained video generators are steered by structure and motion through control adapters~\citep{zhang2023adding,guo2024sparsectrl,wang2024motionctrl}, and by example videos that carry motion~\citep{zhao2024motiondirector,wei2024dream,xu2026disco,ressler2025dismo}, actions~\citep{sun2024videodemo}, physics~\citep{chen2026viper}, or edits~\citep{wang2026vicedit}. Reading a task from one visual example goes back to image analogies~\citep{hertzmann2001image} and visual prompting~\citep{bar2022visual,wang2023images,wang2023context}, and image models now do it with little or no text~\citep{gu2024analogist,li2025visualcloze,chen2025edittransfer}. In-context conditioning places the example in the same token sequence as the target~\citep{huang2024context,fei2024vdictx}, so a clip, an image, or nothing can fill an input slot without a dedicated branch~\citep{jiang2025vace}. Guidance amplifies a condition by contrast with an empty condition~\citep{ho2022classifier}, a weaker model~\citep{karras2024guiding,ahn2024self,hyung2025stg}, a degraded or learned negative~\citep{sadat2025icg,han2026cdg,li2024reneg}, or an appearance that cancels out of a motion~\citep{huang2025videomage}; our negative is the reference itself with its transition replaced by a dissolve, so its scenes cancel and only the effect is amplified. Evaluation scores realism~\citep{unterthiner2018towards}, general quality~\citep{huang2024vbench}, or similarity to the reference~\citep{radford2021learning,zhao2024videoprism,yatim2024spacetime}, and similarity rewards copying the input~\citep{krojer2024aurora}. No existing dataset shows one two-sided transition effect on many endpoint pairs, and transition benchmarks score how clips connect without a reference effect~\citep{yin2026vcbench}.

\paragraph{Transition effect generation.}
Work on transitions connects two given endpoints, but the transition is chosen by the model or by text, never read from a reference clip. Inbetweening bridges two frames~\citep{reda2022film,feng2024timereversal,xing2024tooncrafter,wang2025generative,zhang2026unitemp}, transition generation steers the bridge with text or structure~\citep{chen2023seine,yang2025versatile,kan2026sage}, and video foundation models take a first and a last frame~\citep{wan2025wan,hacohen2026ltx}; several accept clips at both ends~\citep{zhang2024mavin,kan2026sage,jiang2025vace}. Commercial tools pick from a preset library~\citep{shen2022autotransition}, and morphing~\citep{beier1992feature} and transition detection~\citep{zhang1993automatic,soucek2024transnet} predate generative models. The task that does read an effect from a reference is visual effect transfer: it applies the effect of a reference video to a single start image and leaves the end open~\citep{li2025vfxmaster,bian2026video,jones2026tuning,yang2026effectmaker,ma2026easyvfx}. Most of these methods also give the model text about the effect (\refvfx{} makes it optional): a prompt that names it~\citep{li2025vfxmaster,bian2026video,ma2026easyvfx}, a description a multimodal LLM writes from the reference~\citep{yang2026effectmaker}, or a prompt optimized to match the reference~\citep{zhao2026pflow}. Others generate effects from text alone~\citep{mao2026omni,liu2025vfx,yang2026prompt2effect} or copy pixels of the reference into the target~\citep{guo2024magicvfx}. \segue{} receives no text about the effect, reads it from the reference alone, and must arrive at a given end clip.
\section{\segue{}: Isolating and Amplifying the Reference Transition}
\label{sec:overview}

\segue{} addresses \ctt{} through two complementary mechanisms (Figure~\ref{fig:model}): training-side \emph{reference isolation}, which combines neutral prompts with a one-way attention mask so the model reads the effect strictly from the visual reference (\S\ref{sec:framework}), and inference-side \emph{null-reference guidance} to amplify that effect during inference (\S\ref{sec:inference}). 

\subsection{\cttlong{} (\ctt{}) Formulation}
\label{sec:task}

\ctt{} transfers a transition effect from a visual reference clip $V_r$ (connecting scenes $A_r$ and $B_r$) onto target endpoints $A$ and $B$ in a one-shot manner. Given a text prompt $c$, the model generates a video:
\begin{equation}
    V = f_\theta(V_r, A, B, c).
    \label{eq:ctt}
\end{equation}
The output must satisfy three core criteria: (i) \emph{transition fidelity}, faithfully transferring the reference transition to the generated transition; (ii) \emph{endpoint fidelity}, opening seamlessly on $A$ and closing on $B$, preserving their visual content and motion; and (iii) \emph{reference disentanglement}, preventing the reference's original scenes ($A_r, B_r$) from leaking into the output.
Each endpoint flexibly accommodates a video clip, a static image, or nothing ($B = \varnothing$), unifying two-sided \ctt{} and single-sided visual effect transfer within a single architecture.           %
\subsection{Training: Visual Reference Isolation}
\label{sec:framework}

\begin{figure}[t]
  \centering
  \includegraphics[width=\linewidth]{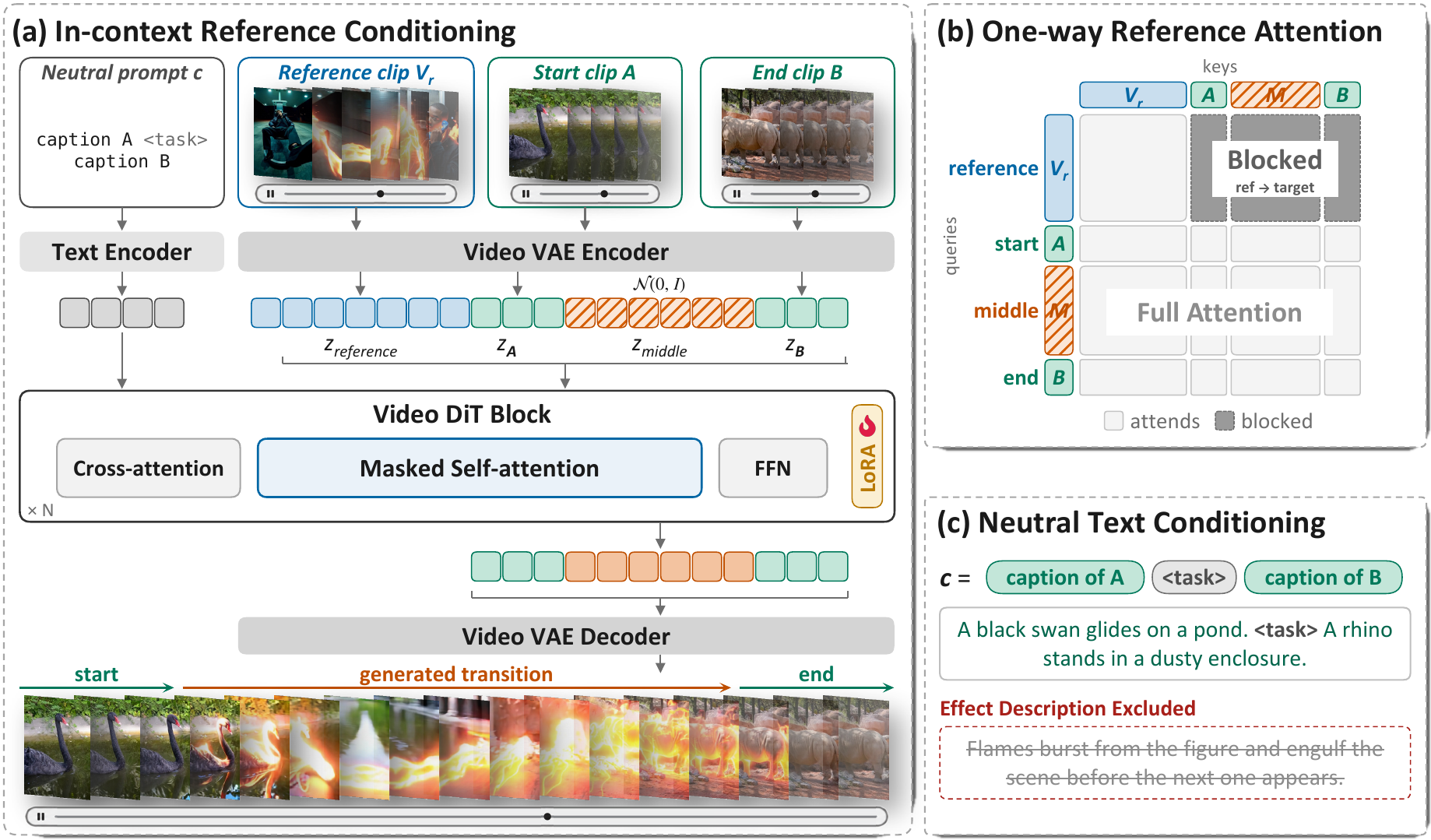}
  \caption{\textbf{\segue{} overview.} (a) In-context reference conditioning: the reference $V_r$, start $A$, and end $B$ are encoded with the noised middle in one sequence, which the LoRA-adapted DiT denoises; the neutral prompt $c$ enters through cross-attention. (b) One-way reference attention: masked self-attention blocks the reference from attending to the target and keeps full attention elsewhere. (c) Neutral text conditioning includes only captions of $A$ and $B$ without the effect description.}
  \label{fig:model}
\end{figure}

We structure training around \emph{visual reference isolation}: the model reads the reference through a neutral prompt and a one-way attention mask, so the prompt says nothing about the effect and the target never rewrites the reference. This closes the text shortcut that previous approaches take (Figure~\ref{fig:motivation}) and leaves the reference as the only input that carries the effect. Below we describe the model (Figure~\ref{fig:model}): its token layout and objective, the attention mask, and the neutral prompt.

\paragraph{In-context reference conditioning.}
Let $V = [A, M, B]$ denote the target video, where the middle $M$ holds the frames that carry the transition between the endpoints $A$ and $B$. Let $\mathbf{z}_r$, $\mathbf{z}_A$, $\mathbf{z}_M$, and $\mathbf{z}_B$ be the VAE latents of the reference, the two endpoints, and the middle. Following in-context conditioning for diffusion transformers~\citep{huang2024context}, we place the reference, the endpoints, and the noised middle in one token sequence on a single positional grid:
\begin{equation}
    x_\sigma = \big[\, \mathbf{z}_r,\; \mathbf{z}_A,\; \mathbf{z}^\sigma_M,\; \mathbf{z}_B \,\big],
    \qquad
    \mathbf{z}^\sigma_M = (1-\sigma)\mathbf{z}_M + \sigma \epsilon,
    \label{eq:seq}
\end{equation}
where $\sigma \in [0, 1]$ is the noise level, and the diffusion transformer reads $x_\sigma$ as one sequence of tokens. Only the middle is noised: $\mathbf{z}_r$, $\mathbf{z}_A$, and $\mathbf{z}_B$ are given without noise and serve as context. This sequential layout handles every endpoint type without a dedicated conditioning branch, in the spirit of unified video conditioning~\citep{jiang2025vace}: a clip contributes all of its frames, an image contributes one frame, and an absent endpoint contributes no tokens. We build the training data with different endpoint settings, so one set of adapter weights covers two-sided transitions between video clips or static images as well as single-sided visual effect transfer from a start alone. We train LoRA adapters~\citep{hu2022lora} on a frozen video diffusion backbone, with the flow-matching objective~\citep{lipman2022flow} restricted to the middle:
\begin{equation}
    \mathcal{L}(\theta) = \mathbb{E}_{\sigma,\,\epsilon}
    \Big[\, \big\lVert\, \mathbf{m} \odot \big( u_\theta(x_\sigma, c) - u \big) \big\rVert^2 \,\Big],
    \qquad
    \mathbf{m}_i = \mathbf{1}\{\, i \in M \,\},
    \label{eq:loss}
\end{equation}
where $u$ is the target velocity, $c$ is the text prompt of Eq.~\eqref{eq:prompt}, and $\mathbf{m}$ masks the loss outside $M$. With fixed endpoints, all training signal is focused on the transition.

\paragraph{One-way attention mask.}
With full self-attention, the reference tokens would also read the target tokens, and layer by layer the representation of the reference would drift toward the target scenes. We therefore make the attention one-way. Let $R$ index the reference tokens and $G$ the remaining tokens, the endpoints and the middle. The mask is (Figure~\ref{fig:model}b):
\begin{equation}
    \text{token } i \text{ attends to token } j \iff i \in G \;\;\text{or}\;\; j \in R.
    \label{eq:mask}
\end{equation}
The target tokens $G$ read the reference $R$, but the reference tokens never read $G$. The representation of the reference therefore depends on $V_r$ alone, in every layer, and the target cannot alter it. This one-way rule is adapted from the reference attention of \vfxmaster{}~\citep{li2025vfxmaster}, which introduced it for effect transfer onto a single frame; their mask also covers text tokens and a reference prompt, neither of which is in our sequence (refer to Appendix~\ref{app:train}).

\paragraph{Neutral text conditioning.}
The text shortcut runs through the prompt: if the prompt named the effect, the model could read it from there and ignore the motion in the reference. We therefore use a neutral prompt that describes only the two scenes (Figure~\ref{fig:model}c):
\begin{equation}
    c = \big[\, \text{caption of } A \,\big]\;
        \langle \text{task} \rangle\;
        \big[\, \text{caption of } B \,\big],
    \label{eq:prompt}
\end{equation}
where $\langle \text{task} \rangle$ is a fixed separator token. No word about the effect enters the prompt during training. This leaves the reference as the only source of the transition effect, a clean visual signal that inference can amplify (\S\ref{sec:inference}).

\subsection{Inference: Null-Reference Guidance}
\label{sec:inference}

Reference isolation makes the reference the carrier of the transition effect. At inference, we want a guidance that amplifies the transition effect without reference scene leakage.

\paragraph{Negative reference.}
Classifier-free guidance~\citep{ho2022classifier} amplifies a condition by steering away from the prediction without it, or away from a negative condition~\citep{ban2024negative}. With the reference as the condition, this reads $\hat{v} = v(\varnothing) + w\,\big(v(V_r) - v(\varnothing)\big)$. But a reference clip carries both, the transition effect and its scenes $(A_r, B_r)$; and an empty reference removes both. The contrast therefore amplifies the scenes of the reference along with its effect, which is the reference scene leakage we want to avoid (Figure~\ref{fig:nullcmp} in Appendix~\ref{app:perop}). Instead, we want a negative reference that keeps the reference's scenes and removes only the effect.

\begin{figure}[t]
  \centering
  \includegraphics[width=\linewidth]{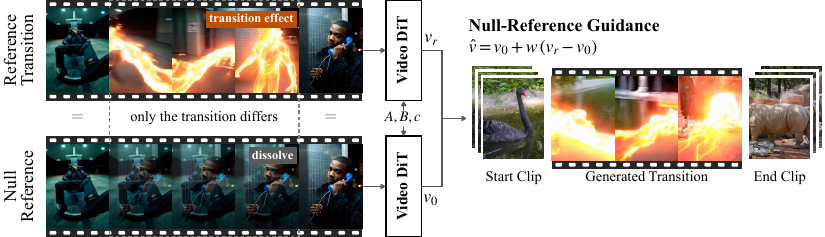}
  \caption{\textbf{Null-reference guidance.} The null reference $V_r^0$ replaces the transition in $V_r$ with a dissolve between the same reference endpoints. The model processes both with identical endpoints $(A,B)$ and prompt $c$, so their predictions, $v_r$ with the reference and $v_0$ with the null reference, differ primarily in the transition effect; guidance steers away from the dissolve and toward the effect.}
  \label{fig:nrg}
\end{figure}

\paragraph{The null transition.}
We observe such a condition in the base model. Given two endpoints and a prompt that describes only the two scenes, \ltx{}~\citep{hacohen2026ltx} mostly produces a straight cut or a dissolve (Appendix~\ref{app:lerp}): it keeps both scenes and adds no motion and no material of its own. We take the dissolve, a linear blend of two frames, as the \emph{null transition},
\begin{equation}
    T_0(a, b):\quad V_\tau = (1 - \tau)\, a + \tau\, b,
    \qquad \tau \in [0, 1],
    \label{eq:null}
\end{equation}
where $a$ is the last frame of the start and $b$ the first frame of the end. The \emph{null reference} is the reference with its own transition replaced by the null transition:
\begin{equation}
    V_r^0 = \big[\, A_r,\; T_0(A_r, B_r),\; B_r \,\big].
    \label{eq:nullref}
\end{equation}
$V_r$ and $V_r^0$ show the same scenes and differ only in the transition, which is the effect in $V_r$ and a dissolve in $V_r^0$ (Figure~\ref{fig:nrg}).

\paragraph{Null-reference guidance (NRG).}
Like classifier-free guidance, at each sampling step we run the model twice, with the same endpoints $(A, B)$, the same prompt $c$, and the same noise, and change only the reference. With $v_r = u_\theta\big(x_\sigma(V_r), c\big)$ and $v_0 = u_\theta\big(x_\sigma(V_r^0), c\big)$,
\begin{equation}
    \hat{v} = v_0 + w\,(v_r - v_0), \qquad w \geq 1.
    \label{eq:nrg}
\end{equation}
At $w = 1$ this is the model without guidance; for $w > 1$ the sample moves away from the dissolve and toward the effect of the reference. Everything except the effect is shared by the two passes, so the contrast is specific to the effect. \rev{As the reference transition approaches a dissolve, the two passes become similar and the guidance term approaches zero. NRG therefore has little effect on dissolve-like transitions, which the base model already produces by default (Appendix~\ref{app:lerp}). We quantify NRG gains across transition types in Table~\ref{tab:nrgtype} (Appendix~\ref{app:ablations}).} The null reference is a valid video and not an empty input, so no condition dropout is needed in training and NRG applies to the trained model as it is.

\subsection{\segue{}-Benchmark}
\label{sec:benchmark}
\ifarxiv\else\begin{wraptable}[\ifarxiv 9\else 13\fi]{r}{0.46\textwidth}   %
    \vspace{-\baselineskip}
    \centering
    \caption{\textbf{Data sources.} A training pair is a target video with up to three references of the same effect. External effects include the 34 EffectData effects, used only for evaluation.}
    \label{tab:sources}
    \scriptsize
    \setlength{\tabcolsep}{2.5pt}
    \resizebox{\linewidth}{!}{%
    \begin{tabular}{lrrrrrr}
        \toprule
        Source & \shortstack{Effect\\types} & \shortstack{Distinct\\endpoints} & Videos & \shortstack{Training\\pairs} & \shortstack{Effects per\\endpoint} & \shortstack{Endpoints\\per effect} \\
        \midrule
        Creative transitions   & 26      & 350     & 1{,}364  & 4{,}060  & 3.6  & 49.0 \\
        Procedural transitions & 1{,}590 & 1{,}103 & 15{,}436 & 46{,}308 & 14.0 & 9.7 \\
        External effects       & 76      & 2{,}051 & 2{,}034  & 6{,}000  & 1.0  & 27.2 \\
        \midrule
        Total                  & 1{,}692 & 3{,}504 & 18{,}834 & 56{,}368 & 5.4  & 11.1 \\
        \bottomrule
    \end{tabular}
    }
    \vspace{-0.5\baselineskip}
\end{wraptable}
\fi   %

Existing video datasets rarely show one transition effect on many scenes: transitions in available footage are one-off instances tied to specific clips. We therefore build the \segue{}-Benchmark, in which every transition effect is a class with many videos. A training sample pairs two videos of the same class on different endpoints: one is the reference $V_r$, and the other is the target $V$, whose endpoints $(A, B)$ are given as conditions. The data is \emph{cross-paired}: each transition effect appears on many endpoint pairs, and each endpoint pair under many transition effects, so effects and endpoints are decorrelated and neither can be inferred from the other.

\ifarxiv\begin{wraptable}[\ifarxiv 9\else 13\fi]{r}{0.46\textwidth}   %
    \vspace{-\baselineskip}
    \centering
    \caption{\textbf{Data sources.} A training pair is a target video with up to three references of the same effect. External effects include the 34 EffectData effects, used only for evaluation.}
    \label{tab:sources}
    \scriptsize
    \setlength{\tabcolsep}{2.5pt}
    \resizebox{\linewidth}{!}{%
    \begin{tabular}{lrrrrrr}
        \toprule
        Source & \shortstack{Effect\\types} & \shortstack{Distinct\\endpoints} & Videos & \shortstack{Training\\pairs} & \shortstack{Effects per\\endpoint} & \shortstack{Endpoints\\per effect} \\
        \midrule
        Creative transitions   & 26      & 350     & 1{,}364  & 4{,}060  & 3.6  & 49.0 \\
        Procedural transitions & 1{,}590 & 1{,}103 & 15{,}436 & 46{,}308 & 14.0 & 9.7 \\
        External effects       & 76      & 2{,}051 & 2{,}034  & 6{,}000  & 1.0  & 27.2 \\
        \midrule
        Total                  & 1{,}692 & 3{,}504 & 18{,}834 & 56{,}368 & 5.4  & 11.1 \\
        \bottomrule
    \end{tabular}
    }
    \vspace{-0.5\baselineskip}
\end{wraptable}
\fi
The training set combines three sources (Table~\ref{tab:sources}), chosen to cover many kinds of transitions and to keep the pairing dense, with many effects per endpoint and many endpoints per effect. (i) \emph{Creative transitions}: 26 preset transition effects whose videos we collect from commercial video generation models, following prior effect datasets~\citep{li2025vfxmaster,yang2026effectmaker}. Each effect is learned by a small LoRA and rendered on a shared set of endpoints, and every video is inspected by hand. (ii) \emph{Procedural transitions}: 1,590 parametric transitions, such as geometric wipes, volumetric zooms, fluid warps, and alpha-blended shaders with randomized direction, speed, and boundary softness, rendered on shared endpoints. (iii) \emph{External visual effects} from prior visual effect transfer work, \refvfx{}~\citep{jones2026tuning} and EffectData~\citep{yang2026effectmaker}, which add single-sided visual effect transfer and other endpoint configurations. In total, the benchmark has 1,692 effect types and 18,834 videos. The 34 EffectData effects and 17 classes of our own sources are held out of training and form the evaluation dataset (Section~\ref{sec:evalsetup}).
\section{Experiments}
\label{sec:exp}

\subsection{Experimental Setup}
\label{sec:evalsetup}

\paragraph{Implementation details.}
\label{sec:impl}
\segue{} is a rank-128 LoRA adapter~\citep{hu2022lora} on the frozen \ltx{} 19B model~\citep{hacohen2026ltx}, trained on 8 H100 GPUs on the 56K training pairs of Table~\ref{tab:sources}. Outputs are $480 \times 640$ at 24\,fps and 121 frames long. We sample with 30 steps and null-reference guidance at $w = 6$ in every table; Appendix~\ref{app:ablations} sweeps $w$ and Appendix~\ref{app:train} gives details.

\begin{table}[t]
    \centering
    \caption{\textbf{Evaluation metrics.} $V$: the generated video; $V_r$: the reference clip; $\mathcal{R}$: real videos of the
    reference's transition class. Full definitions in Appendix~\ref{app:metric}.}
    \label{tab:metricsmain}
    \scriptsize
    \setlength{\tabcolsep}{4pt}
    \renewcommand{\arraystretch}{1.1}
    \newcolumntype{L}[1]{>{\raggedright\arraybackslash}p{#1}}
    \newcolumntype{Y}{>{\raggedright\arraybackslash}X}
    \begin{tabularx}{\linewidth}{L{0.16\linewidth}YL{0.50\linewidth}}
        \toprule
        Metric & What it checks & Score \\
        \midrule
        \rowcolor{black!5}Transition fidelity\,$\uparrow$ & Does $V$ show the transition effect of $V_r$?
            & $100 \cdot \mathrm{sim}(V, \mathcal{R}) \,/\, \mathrm{sim}(\mathcal{R}, \mathcal{R})$; sim: change over time
              (DINOv2), motion (CoTracker3), appearance; real videos score 100 \\
        Motion fidelity\,$\uparrow$ & Does $V$ move like $V_r$?
            & $100 \cdot \mathrm{corr}\big(\mathrm{Tracks}(V), \mathrm{Tracks}(V_r)\big)$, CoTracker3 point tracks \\
        \rowcolor{black!5}Action KL\,$\downarrow$ & Does $V$ show the same action as $V_r$?
            & $\mathrm{KL}\big(\mathrm{Swin}(V_r) \,\|\, \mathrm{Swin}(V)\big)$, Video Swin action probabilities \\
        Reference similarity\,$\uparrow$ & Does $V$ look like $V_r$?
            & $100 \cdot \cos\big(\mathrm{VideoPrism}(V), \mathrm{VideoPrism}(V_r)\big)$ \\
        \rowcolor{black!5}Seam-free rate\,$\uparrow$ & Does $V$ leave the given clips without a visible cut?
            & \% of outputs whose LPIPS change at the hand-off is within their usual frame-to-frame change \\
        \bottomrule
    \end{tabularx}
\end{table}
\paragraph{Evaluation dataset and metrics.}
\label{sec:evaldata}\label{sec:metrics}
The evaluation dataset of \segue{}-Benchmark is fully held out: neither its transitions nor its endpoints appear in training. Each evaluation sample consists of a reference clip that shows a transition effect and the target endpoints to which the effect must be transferred: a start and an end for two-sided \ctt{} (38 samples), or the start alone for single-sided visual effect transfer (183 samples). The 221 samples cover 51 transition classes (the 34 EffectData classes and 17 classes held out from our own sources) and 150 distinct endpoints, and each sample is generated with two seeds. Appendix~\ref{app:grid} lists the classes and the samples. We report five metrics, defined in Table~\ref{tab:metricsmain}. \emph{Transition fidelity} compares the output with real videos of the reference's transition class in change over time, motion, and appearance, scaled so that real videos score 100. \emph{Motion fidelity}, \emph{action KL}, and \emph{reference similarity} compare the output with the reference clip in tracked motion, action class, and appearance, and the \emph{seam-free rate} checks the hand-off to the given clips for a visible cut. Every system is run with two prompts: the neutral prompt (N), which describes only the endpoints, as in Eq.~\eqref{eq:prompt}, and the effect prompt (E), which also describes the reference transition. For every metric we report both scores and $|\Delta|$, the absolute difference between them. A small $|\Delta|$ means that the model reads the transition mainly from the reference and depends little on the text prompt.

\paragraph{Baselines.}
\label{sec:baselines}
No previous approach is designed for a reference clip together with two endpoints. For two-sided \ctt{}, we therefore adapt existing models as baselines: the base \ltx{}~\citep{hacohen2026ltx}, VACE~\citep{jiang2025vace}, and Wan2.1 FLF2V~\citep{wan2025wan} take both endpoints but no reference, so the effect prompt is their only source of the effect; \refvfx{}~\citep{jones2026tuning} takes the reference and the start endpoint, and we adapt it by giving it the end endpoint as well. For single-sided visual effect transfer, we compare with the previous visual effect transfer methods \vap{}~\citep{bian2026video}, \vfxmaster{}~\citep{li2025vfxmaster}, and \refvfx{}, which take the reference and the start endpoint. Each runs in its native setting, with its own prompt format, resolution, and frame count.

\subsection{Analysis}
\label{sec:analysis}

\begin{table}[t!]
    \centering
    \caption{\textbf{Comparison with previous approaches and adapted baselines} on the evaluation dataset. N and E: the neutral prompt, which describes only the endpoints, and the effect prompt, which also describes the reference transition; $|\Delta|$: the absolute gap between them. \segue{} w/o NRG: no guidance ($w = 1$), not ranked; small numbers under its values: its difference from \segue{} (green: better, red: worse).}
    \label{tab:main}
    \newcommand{\dgood}[1]{{\tiny\textcolor[HTML]{2E7D32}{#1}}}
    \newcommand{\dbad}[1]{{\tiny\textcolor[HTML]{B22222}{#1}}}
    \newcommand{\dsame}[1]{{\tiny}}
    \newcommand{\tabdash}{\noalign{\vskip 1.2pt\hbox to \linewidth{\color{black!60}\leaders\hbox to 3pt{\hss\vrule height 0.4pt width 1.6pt\hss}\hfill}\vskip 1.2pt}}
    \scriptsize
    \setlength{\tabcolsep}{2pt}
    \begin{tabular*}{\linewidth}{@{\extracolsep{\fill}}lccc!{\hspace{3pt}}ccc!{\hspace{3pt}}ccc!{\hspace{3pt}}ccc!{\hspace{3pt}}ccc@{}}
        \toprule
        & \multicolumn{3}{c}{Transition fidelity\,$\uparrow$} & \multicolumn{3}{c}{Motion fid.\,($\times$100)\,$\uparrow$} & \multicolumn{3}{c}{Action KL\,$\downarrow$} & \multicolumn{3}{c}{Ref sim.\,($\times$100)\,$\uparrow$} & \multicolumn{3}{c}{Seam-free\,\%\,$\uparrow$} \\
        \cmidrule(lr){2-4}\cmidrule(lr){5-7}\cmidrule(lr){8-10}\cmidrule(lr){11-13}\cmidrule(lr){14-16}
         & N & E & $|\Delta|$ & N & E & $|\Delta|$ & N & E & $|\Delta|$ & N & E & $|\Delta|$ & N & E & $|\Delta|$ \\
        \hline
        \tabsec{16}{Transition effect generation: both endpoints given} \\
        Base \ltx{} & 49.8 & 84.7 & 34.9 & 8.8 & 8.1 & 0.7 & 5.98 & 5.48 & 0.50 & 94.8 & 95.3 & 0.5 & 80.3 & 93.4 & 13.1 \\
        VACE & 51.9 & 77.0 & 25.1 & 11.9 & 9.9 & 2.0 & 5.71 & 5.04 & 0.67 & 95.1 & 95.2 & \textbf{0.1} & 92.1 & 93.4 & \textbf{1.3} \\
        \refvfx{} & 49.8 & 70.5 & 20.7 & \textbf{14.1} & 11.4 & 2.7 & 5.68 & 5.38 & 0.30 & 95.2 & 95.4 & 0.2 & 88.2 & 97.4 & 9.2 \\
        Wan2.1 FLF2V & 56.6 & 84.3 & 27.7 & 9.4 & 7.4 & 2.0 & 5.59 & 4.98 & 0.61 & 95.0 & 95.2 & 0.2 & 94.7 & \textbf{100.0} & 5.3 \\
        \addlinespace[2pt]
        \segue{} & \textbf{97.0} & \textbf{97.9} & \textbf{0.9} & 12.8 & \textbf{12.5} & \textbf{0.3} & \textbf{4.62} & \textbf{4.52} & \textbf{0.10} & \textbf{95.9} & \textbf{96.0} & \textbf{0.1} & \textbf{97.4} & 98.7 & \textbf{1.3} \\
        \tabdash
        \segue{} w/o NRG & 92.9 & 96.2 & 3.3 & 11.7 & 12.0 & 0.3 & 5.03 & 5.14 & 0.11 & 95.8 & 95.9 & 0.1 & 100.0 & 100.0 & 0.0 \\[-2.5pt]
         & \dbad{$-$4.1} & \dbad{$-$1.7} & \dbad{+2.4} & \dbad{$-$1.1} & \dbad{$-$0.5} & \dsame{$\pm$0.0} & \dbad{+0.41} & \dbad{+0.62} & \dbad{+0.01} & \dbad{$-$0.1} & \dbad{$-$0.1} & \dsame{$\pm$0.0} & \dgood{+2.6} & \dgood{+1.3} & \dgood{$-$1.3} \\
        \hline
        \tabsec{16}{Visual effect transfer: start endpoint given} \\
        \vap{} & 40.9 & 70.3 & 29.4 & 12.9 & 13.7 & 0.8 & 5.69 & 4.41 & 1.28 & 94.7 & 95.8 & 1.1 & 92.1 & 93.2 & 1.1 \\
        \vfxmaster{} & 49.9 & 85.5 & 35.6 & 17.5 & 19.2 & 1.7 & 5.48 & 4.50 & 0.98 & 94.9 & 96.2 & 1.3 & \textbf{96.7} & \textbf{99.7} & 3.0 \\
        \refvfx{} & 42.1 & 66.3 & 24.2 & 18.0 & 16.1 & 1.9 & 5.60 & 4.75 & 0.85 & 94.9 & 95.7 & 0.8 & 66.7 & 94.5 & 27.8 \\
        \addlinespace[2pt]
        \segue{} & \textbf{92.6} & \textbf{94.0} & \textbf{1.4} & \textbf{22.6} & \textbf{22.7} & \textbf{0.1} & \textbf{4.15} & \textbf{4.05} & \textbf{0.10} & \textbf{96.5} & \textbf{96.6} & \textbf{0.1} & 93.4 & 92.9 & \textbf{0.5} \\
        \tabdash
        \segue{} w/o NRG & 86.4 & 91.0 & 4.6 & 20.3 & 20.4 & 0.1 & 4.52 & 4.29 & 0.23 & 96.3 & 96.5 & 0.2 & 96.4 & 96.2 & 0.2 \\[-2.5pt]
         & \dbad{$-$6.2} & \dbad{$-$3.0} & \dbad{+3.2} & \dbad{$-$2.3} & \dbad{$-$2.3} & \dsame{$\pm$0.0} & \dbad{+0.37} & \dbad{+0.24} & \dbad{+0.13} & \dbad{$-$0.2} & \dbad{$-$0.1} & \dbad{+0.1} & \dgood{+3.0} & \dgood{+3.3} & \dgood{$-$0.3} \\
        \bottomrule
    \end{tabular*}
\end{table}

\begin{figure}[t]
  \centering
  \begin{minipage}[t]{0.5\linewidth}
    \vspace{0pt}
    \centering
    \includegraphics[width=\linewidth]{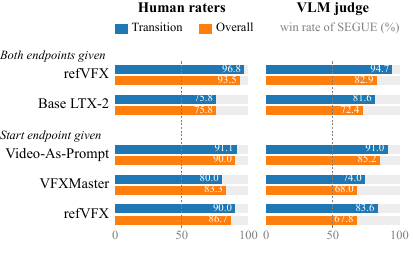}
    \caption{\textbf{Human study and VLM judge.} Win rate (\%) against each system for transition and overall quality. Protocol in Appendix~\ref{app:human}.}
    \label{fig:study}
  \end{minipage}\hfill
  \begin{minipage}[t]{0.46\linewidth}
    \vspace{0pt}
    \centering
    \captionof{table}{\textbf{Flagged clips} (\%) by the raters' majority and by the VLM judge, for a missing
    effect, a missed endpoint (End.; for the judge, among clips with an effect), and leaked reference content.}
    \label{tab:study}
    \scriptsize
    \setlength{\tabcolsep}{1.5pt}
    \begin{tabular*}{\linewidth}{@{\extracolsep{\fill}}lcccccc@{}}
    \toprule
    & \multicolumn{3}{c}{Raters (\%)\,$\downarrow$} & \multicolumn{3}{c}{VLM judge (\%)\,$\downarrow$} \\
    \cmidrule(lr){2-4}\cmidrule(lr){5-7}
    & Effect & End. & Leak & Effect & End. & Leak \\
    \hline
    \tabsec{7}{Both endpoints given} \\
    \refvfx{} & 40.3 & 4.8 & 19.4 & 5.3 & 8.3 & 0.0 \\
    Base \ltx{} & 40.3 & 0.0 & 6.5 & 10.5 & 2.9 & -- \\
    \segue{} & 0.0 & 0.0 & 4.8 & 0.0 & 5.3 & 5.3 \\
    \hline
    \tabsec{7}{Start endpoint given} \\
    \vap{} & 32.2 & 10.0 & 23.3 & 4.4 & 0.0 & 1.1 \\
    \vfxmaster{} & 24.4 & 5.6 & 14.4 & 0.0 & 0.0 & 0.0 \\
    \refvfx{} & 26.7 & 3.3 & 16.7 & 1.1 & 0.6 & 0.0 \\
    \segue{} & 0.0 & 0.0 & 1.1 & 2.2 & 0.0 & 0.0 \\
    \bottomrule
    \end{tabular*}
  \end{minipage}
\end{figure}

\begin{figure}[!htb]
  \centering
  \includegraphics[width=\linewidth]{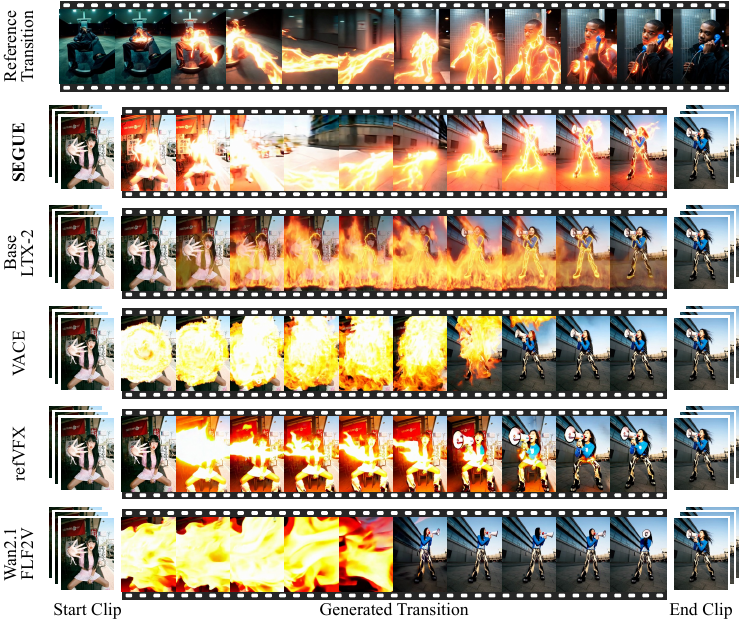}   %
  \caption{\textbf{Qualitative comparison}, both endpoints given. Top: the reference; below: one system per row, frames in time order. \ifarxiv Videos and additional results are on the project page.\else Videos and additional results are available in the supplemental.\fi}
  \label{fig:qualitative}
\end{figure}

\paragraph{\segue{} transitions are more faithful.}
\label{sec:quant}
To test whether a model can transfer a transition effect it has never seen, we compare \segue{} with previous approaches on held-out effects, in both tasks with the neutral and the effect prompt (Table~\ref{tab:main}). \segue{} produces the most faithful transitions in both tasks: with the neutral prompt, it scores close to real videos of the same effect, while previous approaches reach about half of that. Even with the effect prompt, they stay below \segue{}, which shows that a text description cannot convey the motion and timing that the reference clip shows. The other reference metrics agree, and the seam-free rate shows that the hand-off to the given clips stays clean. The qualitative comparison makes the difference visible (Figure~\ref{fig:qualitative}; Figure~\ref{fig:qualitativevet} in Appendix~\ref{app:perop} for the start alone): \segue{} reproduces the dynamics, timing, and appearance of the reference between the endpoints without leaking its scenes, whereas previous approaches render a generic frame-filling effect even with the reference or its description.

\paragraph{\segue{} uses the visual signal of the reference transition.}
\label{sec:textdep}
Our hypothesis is that previous approaches take the effect from the text prompt rather than from the reference. To measure this, we run every system with the effect prompt (E) and with the neutral prompt (N) and report the gap $|\Delta|$ (Table~\ref{tab:main}). Previous approaches lose most of the effect once the prompt stops describing it, so their visual reference contributes little. \segue{} barely changes on any metric: it reads the transition from the reference clip, and the prompt adds almost nothing to what the reference already carries.

\paragraph{Human study and VLM judge.}
\label{sec:human}
Automatic metrics may miss how people perceive a transition, so we also ask human raters and a VLM judge to choose between \segue{} and one other model, on the transition and overall, and to flag clips with a missing effect, missed endpoint, or leaked reference content (Figure~\ref{fig:study} and Table~\ref{tab:study}; protocol in Appendix~\ref{app:human}). With both endpoints given, the opponents are \refvfx{} and the base \ltx{} with the effect prompt; with the start alone, \vap{}, \vfxmaster{}, and \refvfx{}. Raters and the judge prefer \segue{} over every other model, significantly on both transition fidelity and overall quality. The flags follow the same order; the judge is more lenient than the raters on whether an effect is present.

\paragraph{\rev{Reliability of transition fidelity.}}
\rev{Transition fidelity (TF) achieves the highest agreement with human preference (86.7\%) and VLM judgments (87.3\%) among the evaluated metrics. In separate retrieval and separation tests, without metric training or tuning, it also achieves the highest effect-retrieval accuracy (73\%) and the second-highest separation. These results support its ability to capture transition effects. However, TF does not measure reference copying, smoothness at endpoint boundaries or overall naturalness. It should therefore be evaluated alongside complementary metrics and human judgments. See Appendix~\ref{app:metric}, Tables~\ref{tab:metrichuman}--\ref{tab:retrieval}, for the evaluation setup and full results.}

\paragraph{\rev{Ablation studies.}}
\label{sec:ablations}
\rev{We first isolate the effects of neutral-prompt training and one-way attention (Table~\ref{tab:ablationpromptattn}). Neutral prompts produce the largest reduction in text dependence, roughly halving the fidelity gap between neutral and effect prompts. One-way attention reduces this gap further, and combining both gives the highest neutral-prompt fidelity and smallest gap. With full training, the gap falls to 3.6, and NRG further reduces it to 1.7. We then evaluate NRG across guidance strengths (Table~\ref{tab:weight}), transition types (Table~\ref{tab:nrgtype}) and alternative negative references (Table~\ref{tab:nulldesigns}). NRG improves creative and EffectData effects, while dissolve-like targets already perform well without guidance. All tested negatives improve effect fidelity, but the dissolve gives the smoothest endpoint boundaries and least reference copying among the guided variants. See Appendix~\ref{app:ablations} for the guidance sweep and full ablation results.}
\section{Conclusion}
\label{sec:conclusion}

We introduced \cttlong{} (\ctt{}) and \segue{}, a model that transfers a transition effect from a single reference clip to new endpoints, whether video clips, static images, or absent. \segue{} combines visual reference isolation during training with null-reference guidance at inference, so the transition is read and amplified from the visual demonstration rather than from the effect prompt. We also introduced the \segue{}-Benchmark, a cross-paired dataset with a held-out evaluation dataset and the transition fidelity metric. Our results show that a visual reference alone is sufficient to specify and reuse a transition effect across endpoints.

\paragraph{Limitations.}
Two failure modes appear on a few samples, shown in Figure~\ref{fig:limitations} of Appendix~\ref{app:perop}. As the guidance weight grows, on some samples we observed scene leakage from the reference, and for one specific endpoint that shows intense motion, the endpoints were not preserved. As any instance of classifier-free guidance, our null-reference guidance doubles the model evaluations per sampling step, and all results are on one base model, \ltx{}~\citep{hacohen2026ltx}.

\ifarxiv
\subsubsection*{Acknowledgements}
We acknowledge the support of the Health Care Engineering Systems Center within the Coordinated Science Lab at UIUC.
\fi
\ifarxiv\else
\clearpage
\input{statements/ai_use}            %
\input{statements/ethics}            %
\input{statements/reproducibility}   %
\fi

\clearpage
\bibliography{references}
\bibliographystyle{iclr2027_conference}

\beginsupplementary
\section{Dataset construction}
\label{app:data}

\paragraph{Creative transitions.}
Following prior effect datasets~\citep{li2025vfxmaster,yang2026effectmaker}, we collect videos of 43 preset transition effects from commercial video generation models: 26 effects for training and 17 held out. We standardise every clip to 121 frames at 24\,fps and $480 \times 640$ pixels. To show one transition on many scenes, we train a small LoRA per training transition on its collected clips and render the transition onto 350 shared endpoints: the 139 collected scenes and 211 clips from DAVIS~\citep{pont20172017} and HumanVid~\citep{wang2024humanvid}. Every rendered clip is inspected by hand, and clips with a boundary artifact, a frozen or black frame, or a changed start scene are removed; 192 of 1,417 rendered clips were rejected. The row of Table~\ref{tab:sources} counts the 139 collected clips and the 1,225 kept renders.

\paragraph{Procedural transitions.}
Each procedural transition is a GLSL shader from the gl-transitions bank~\citep{gltransitions} with sampled parameters: the shader uniforms, an easing curve, the onset and release times, a flip, and a swap of the two inputs. The timing is part of the transition, so two samplings of one shader are two transitions. We render each transition between 10 pairs of real clips at 121 frames, 24\,fps, and $480 \times 640$; the first and last 9 frames are the source clips byte for byte, so the endpoints are exact by construction. The endpoint clips come from DAVIS, HumanVid, and VC-Bench~\citep{yin2026vcbench}. Shader families are held out whole: 10 families are never rendered, each with a same-genre cousin in training.

\paragraph{External effects.}
From \refvfx{}~\citep{jones2026tuning} we take all 42 effect LoRAs at their native 33 frames, 16\,fps, and $832 \times 448$, one effect per clip and 47 to 50 clips per effect; 5 further \refvfx{} effects are held out. From EffectData~\citep{yang2026effectmaker} we select 34 effects as 17 couples that share a start subject, keep one reference video per effect and 51 start subjects, and never train on them.

\paragraph{Training pairs.}
Within each effect the clips are sorted, and every clip is the target of up to three pairs whose references are the next three clips of the same effect in cyclic order. The reference and the target never share an endpoint, and effects with fewer than two clips are dropped. This gives the 56,368 pairs of Table~\ref{tab:sources} from 18,800 videos.

\paragraph{Prompts.}
Every training prompt is rendered from per-clip scene descriptions by one template: the start scene, the trigger token, and, for a two-sided clip, the end scene, as in Eq.~\eqref{eq:prompt}. The descriptions are written from the first nine frames of the clip, and for a two-sided clip also from the last nine, so no prompt describes the transition. A leak check confirms that a judge given only the prompt cannot identify the effect above chance.

\paragraph{Held-out sets.}
The held-out class tier of the evaluation dataset uses the 17 held-out creative transitions and the 34 EffectData effects. Its 150 endpoints and 68 reference clips are disjoint from the training set, including the DAVIS clips that the creative renders use.
\section{Training details}
\label{app:train}
\paragraph{Adapter.}
\segue{} is a rank-128 LoRA on the frozen \ltx{} 19B model. The adapter sits on the query, key, value, and output projections of self- and cross-attention and on both feed-forward projections of every block, 654M parameters, 3.4\% of the backbone. The backbone, its VAE, and its text encoder stay frozen.

\paragraph{Sequence layout.}
The token sequence of Eq.~\eqref{eq:seq} holds the reference in full, followed by the target. A given start clip contributes its first 9 frames and a given end clip its last 8, a static image contributes one frame, and an absent endpoint contributes no tokens. The middle fills the rest of a 121-frame target, so a transition between two clips has 104 generated frames. All videos are $480 \times 640$ at 24\,fps; the EffectData samples use 81-frame targets, the length of their source clips. The reference and the endpoints enter clean, only the middle is noised, and the loss covers only the middle (Eq.~\eqref{eq:loss}).

\paragraph{Optimization.}
We train with AdamW at a learning rate of $10^{-4}$, decayed linearly to $10^{-5}$ over 10,000 steps without warmup, in bf16 with an effective batch of 8 (8 H100 GPUs, one sample each), about 3 seconds per step. The endpoint setting is chosen per sample from the pairs of Table~\ref{tab:sources}, so one adapter serves clips, images, and absent endpoints.

\paragraph{Sampling.}
Sampling uses 30 steps, the backbone's default text guidance of 4 and spatio-temporal guidance of 1, and null-reference guidance at $w = 6$ (Section~\ref{sec:inference}). Null-reference guidance runs the model twice per step, once with the reference and once with the null reference, so a clip costs twice the time of the base model.

\paragraph{One-way reference attention.}
The rule of Eq.~\eqref{eq:mask} is that of the reference attention of \vfxmaster{}~\citep{li2025vfxmaster}, applied to the video tokens: the reference tokens attend only to themselves, and the endpoint and middle tokens attend to everything. The mask of \vfxmaster{} also covers the text tokens of its sequence and a reference prompt; \ltx{} reads the text through cross-attention and \segue{} uses no reference prompt, so no text token is masked. We implement the rule without a mask tensor, as two unmasked attention calls per layer, one for the reference queries over the reference keys and one for the target queries over all keys, and verified that the result matches the dense-mask attention. Both calls run on the fused scaled-dot-product attention kernels of PyTorch 2.9.1. For a forward and backward pass over 9,600 tokens in bf16 on an L40S GPU, attention takes 27.0\,ms bidirectional, 82.5\,ms with a dense mask, and 21.8\,ms with the two calls, so the one-way rule is no slower than full attention.
\section{Evaluation metrics}
\label{app:metric}
This appendix defines the five metrics of the result tables, summarized in Table~\ref{tab:metricsmain}. Every metric compares the generated video $V$ with the reference clip $V_r$, or with real videos of the transition effect that $V_r$ shows. All are computed on the whole output at its native resolution and frame rate. Four of them sample the video at fixed fractions of its duration, so clips of different length and frame rate are compared on equal footing; the seam-free rate looks at consecutive frames instead.

\paragraph{Transition fidelity.}
The score asks whether $V$ shows the same transition effect as the real videos of the reference's class, in five steps.
\begin{enumerate}[leftmargin=1.5em, itemsep=1pt, topsep=2pt]
    \item \textbf{Features.} Every frame is described by its DINOv2 class token~\citep{oquab2023dinov2}. The given windows at the two ends of the video, 9/121 and 8/121 of its length, give two endpoint anchors, and the frames between them are resampled to 32 steps of normalized time.
    \item \textbf{Residual.} Each frame feature is projected off the plane spanned by the two anchors. What remains, the residual $r_t$, is the part of the frame that the endpoints do not explain: the effect.
    \item \textbf{Three distances} between $V$ and a real video $U$.
        \emph{Change over time}: the residual steps $r_{t+1} - r_t$ are unit directions, each with a mass equal to its length and a time stamp equal to how far the clip has progressed from the start scene to the end scene; the two sets of steps are matched by optimal transport with cost $(1 - \cos) + 0.25\,|\Delta\text{progress}|$, so a mistimed or differently shaped change costs more.
        \emph{Appearance}: every unit residual frame is matched to its most similar frame in the other video, in both directions and without regard to order; the distance is one minus the mean best cosine.
        \emph{Motion}: CoTracker3~\citep{karaev2024cotracker3} tracks a $20 \times 20$ grid of points; the tracks are smoothed and resampled to 32 steps, and each step is summarized by 18 numbers (fraction of moving points, mean speed, an 8-bin direction histogram weighted by speed, the divergence, curl, and two shear terms of an affine fit to the motion field, the mean translation, and the spread of the moving points), z-scored on the corpus; the distance is one minus the cosine of the two concatenated descriptors, so the order of the steps matters.
    \item \textbf{Normalization.} Each distance becomes a rank: one minus its empirical CDF over all pairs among the 222 real transition clips of our sources. The three ranks are averaged with equal weights, and the average is ranked once more against the same average over the corpus pairs. This puts the three distances on one scale and makes the score comparable across effects.
    \item \textbf{Score.} The similarity of $V$ is averaged over the real videos of the reference's transition class, without the reference itself, and divided by the mean similarity of those real videos to one another. \rev{For an EffectData effect, these are eight other clips of the same effect.} The result is multiplied by 100 and capped at 100, so a real video of the effect scores 100 on average.
\end{enumerate}

\paragraph{Motion fidelity.}
As in \citet{yatim2024spacetime}: the CoTracker3 tracks of $V$ and of $V_r$ are reduced to unit velocity directions at 64 steps; every moving track is matched to the track of the other video with the highest direction correlation, and the mean over both directions is the score. A clip in which no point moves has no score and is left out of the mean.

\paragraph{Action KL.}
Thirty-two uniformly sampled frames of the whole video pass through Video Swin-B~\citep{liu2022videoswin} trained on Kinetics-400. The score is the KL divergence from the class distribution of $V_r$ to that of $V$: it is lowest when the output triggers the same action classes as the reference. The 400 classes serve as a signature of the motion, not as names of the effects.

\paragraph{Reference similarity.}
Cosine similarity between the VideoPrism~\citep{zhao2024videoprism} embeddings of $V$ and of $V_r$, computed with the scorer released with \refvfx{}, so the column is comparable with that paper.

\paragraph{Seam-free rate.}
LPIPS between consecutive frames, $d_t$, is computed over the whole output. At the hand-off from the given start frames to the generated frames, and, when both endpoints are given, at the hand-off back to the given end frames, the step $d_t$ is compared with the other steps of the same video by a robust z-score (median and median absolute deviation). An output is seam-free when the larger of its hand-off z-scores is at most 3. The rate is the share of seam-free outputs.

\paragraph{Reporting.}
Each number in a result table is the mean over the videos of its block: 38 two-endpoint samples with two seeds each (76 videos) and 183 start-only samples with two seeds each (366 videos) on the held-out class tier; our own variants are also reported on the other tiers in Appendix~\ref{app:ablations}. Motion fidelity and reference similarity are printed times 100. $|\Delta|$ is the absolute difference between the two printed means, E and N. Bold marks the best value of each column within a block, separately for N and for E, and the smallest $|\Delta|$.

\paragraph{\rev{Agreement with human judgments.}}\label{app:metrichuman}
\rev{Table~\ref{tab:metrichuman} checks each metric against the human study (Appendix~\ref{app:human}): on every pair, does
the metric prefer the same output as the raters' majority, and as the VLM judge? Transition fidelity agrees most often with
both, 86.7\% with the raters and 87.3\% with the judge, close to the 89.9\% at which the judge and the raters agree with each
other. It does not measure copying of the reference, seams at the hand-off, or the naturalness of the edit; the other
metrics and the human study cover those.}

\begin{table}[H]
    \revfloat
    \centering
    \caption{\textbf{Agreement of each metric with the raters and the VLM judge}: share of the 197 study pairs on which the metric prefers the same output (ties excluded).}
    \label{tab:metrichuman}
    \footnotesize
    \setlength{\tabcolsep}{4pt}
    \begin{tabular}{lcccccc}
        \toprule
        & \multicolumn{3}{c}{Raters (\%)} & \multicolumn{3}{c}{VLM judge (\%)} \\
        \cmidrule(lr){2-4}\cmidrule(lr){5-7}
        Metric & All & Two-sided & Single-sided & All & Two-sided & Single-sided \\
        \midrule
        Transition fidelity & \textbf{86.7} & \textbf{89.5} & \textbf{85.3} & \textbf{87.3} & \textbf{92.7} & \textbf{84.7} \\
        Reference similarity & 81.9 & 80.7 & 82.5 & 85.1 & 87.3 & 84.0 \\
        Motion fidelity & 70.4 & 66.7 & 72.4 & 71.3 & 67.9 & 73.1 \\
        Action KL & 59.3 & 59.6 & 59.2 & 60.9 & 65.5 & 58.8 \\
        \bottomrule
    \end{tabular}
\end{table}

\paragraph{\rev{Effect retrieval.}}\label{app:retrieval}
\rev{Table~\ref{tab:retrieval} asks whether a metric recognizes an effect across different videos. On the 222 real
transition videos of our sources (39 effects), we find each video's closest other video under the metric and check whether
it shows the same effect, and we measure how clearly the metric separates same-effect from different-effect pairs.
Transition fidelity retrieves the right effect most often (73\%) and separates the effects second best, after reference
similarity.}

\begin{table}[H]
    \revfloat
    \centering
    \caption{\textbf{Effect retrieval and separation} on 222 real transition videos (39 effects); separation is Cohen's $d$.}
    \label{tab:retrieval}
    \footnotesize
    \setlength{\tabcolsep}{6pt}
    \begin{tabular}{lcc}
        \toprule
        Metric & Retrieval (\%)\,$\uparrow$ & Separation ($d$)\,$\uparrow$ \\
        \midrule
        Transition fidelity & \textbf{73} & 1.39 \\
        Reference similarity & 64 & \textbf{1.54} \\
        Action KL & 32 & 0.60 \\
        Motion fidelity & 8 & 0.43 \\
        \bottomrule
    \end{tabular}
\end{table}

\section{Human study and VLM judge protocol}
\label{app:human}

\paragraph{Samples and pairs.} A pair shows the reference clip, the given start frame (and the given end frame for transition effect generation), and two outputs on the same sample: \segue{} and one other system, each at its native frame count and resolution. Samples come from the held-out class tier of the evaluation dataset (38 two-sided and 183 single-sided samples). The VLM judge rates every sample of the tier: the 38 two-sided samples against \refvfx{} and Base \ltx{} (76 pairs), and the 183 single-sided samples against \vap{}, \vfxmaster{} and \refvfx{} (549 pairs); seed 42 throughout. The human study rates 197 of these pairs, on 76 samples chosen at random from the tier: 31 two-sided samples (62 pairs) and 45 single-sided samples, one per effect class (135 pairs).

\begin{figure}[t]
  \centering
  \includegraphics[width=\linewidth]{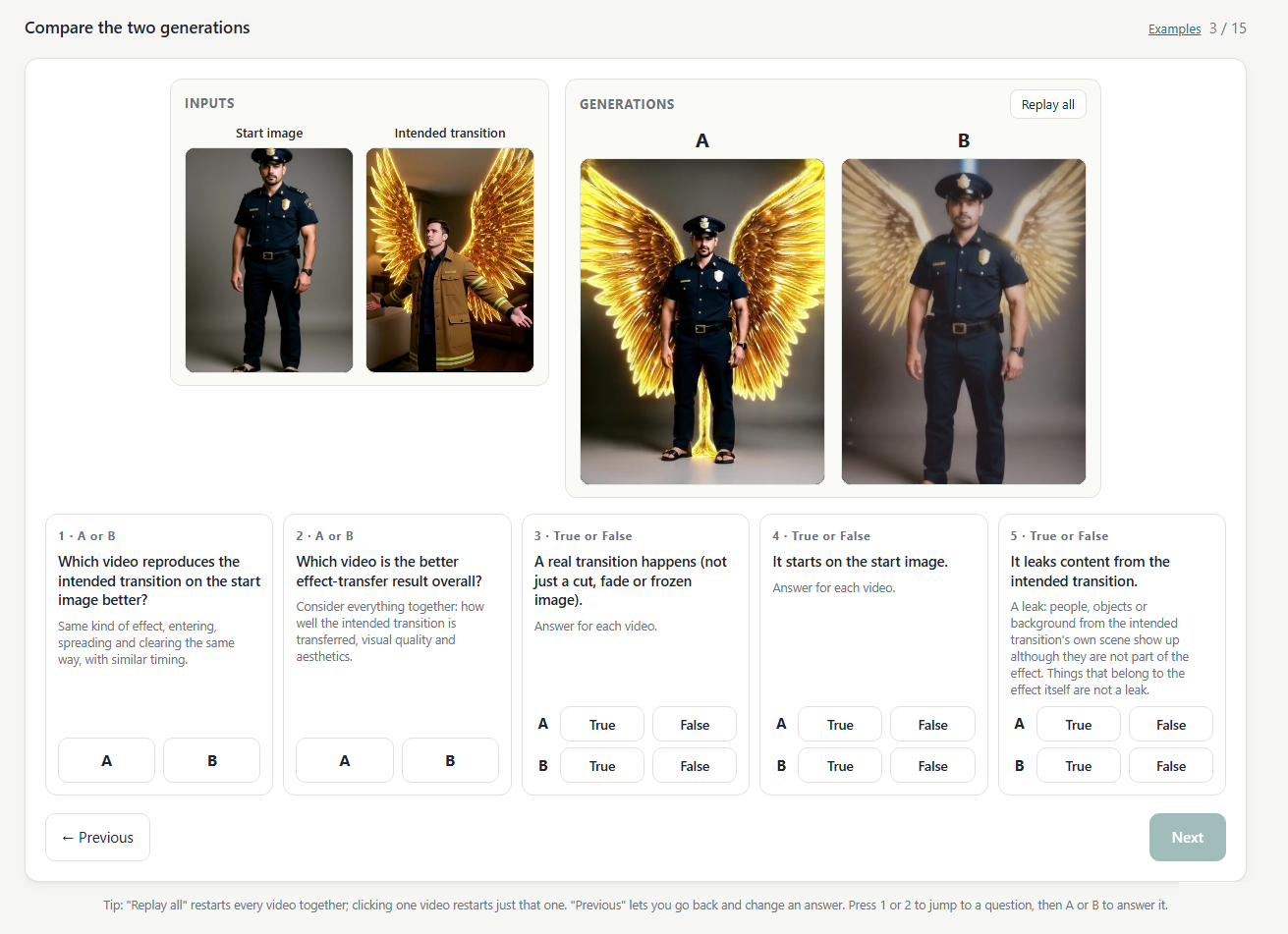}
  \caption{\textbf{The human study page} (a single-sided pair). Inputs on the left (start image, and the end image when given, then the intended transition), the two outputs on the right, and the five questions in one row: two A/B choices and three True/False statements answered for each clip. The questions are reproduced verbatim below.}
  \label{fig:study_page}
\end{figure}

\paragraph{Human raters.} Raters were recruited on Prolific and the study is anonymous: the page asks for no name or identifier, and only the answers are recorded. A rater first reads a short instruction screen, then walks through four worked examples with their answers, and then rates a set of 15 pairs, about 8 minutes. Each pair is shown as in Figure~\ref{fig:study_page}: on the left the given start image (and the given end image for transition effect generation) and the reference, labelled ``intended transition''; on the right the two outputs, labelled A and B, with the side of \segue{} drawn at random per pair. Every clip plays at its own native frame rate and length, holds its last frame when it ends, and all clips of a pair restart together; the instructions state that a shorter or less smooth video is never a reason to prefer one. Five questions follow in one row. Two are forced choices between A and B with no tie: (1) which video reproduces the intended transition on the given image(s) better, and (2) which video is the better effect-transfer result overall, considering the transfer of the intended transition, visual quality and aesthetics. Three are statements answered True or False for A and for B separately: (3) a real transition happens, not just a cut, fade or frozen image; (4) it starts on the start image (and ends on the end image); (5) it leaks content from the intended transition, where a leak is people, objects or background of the reference's own scene that are not part of the effect. A rater may go back to an earlier pair of the set and change an answer; the last answer counts. Pairs are served least-rated first, so every pair reaches three raters before any pair gets a fourth. Nothing on the page names a system, effect or prompt, and media files carry hashed names.

\paragraph{Scoring.} A rater's own answer to statement (3) gates that rater's two choices on the pair: if only one of the two clips shows a transition, that clip wins both choices; if neither does, the pair is a tie for that rater; otherwise the rater's choices stand. Per pair, the gated votes are combined by majority (an even split is a tie), and the win rate of \segue{} against a system is the mean over that system's pairs with a tie counted as half. Significance is a two-sided sign test on the decided pairs; the interval is a Wilson 95\% interval over pairs. Flag rates are per clip: a clip is flagged when the majority of its raters answer False to statement (3) or (4), or True to statement (5); each flag is counted on its own, over all clips of the system. In total, 29 raters gave 529 ratings over the 197 pairs, 2.7 per pair on average. Raters agreed pairwise on 73\% of the gated transition choices, 71\% of the overall choices, and on 74\%, 88\% and 73\% of statements (3), (4) and (5).

\paragraph{Questions as shown to raters.} Verbatim from the page (single-sided wording; for pairs with both endpoints given, question 1 reads ``on the start and end images'' and statement 4 reads ``It starts on the start image and ends on the end image.''). Each question carries a one-line note in smaller type, given here in italics.
\begin{enumerate}[leftmargin=1.6em,itemsep=2pt]
  \item \textit{A or B.} Which video reproduces the intended transition on the start image better? \\ \textit{Same kind of effect, entering, spreading and clearing the same way, with similar timing.}
  \item \textit{A or B.} Which video is the better effect-transfer result overall? \\ \textit{Consider everything together: how well the intended transition is transferred, visual quality and aesthetics.}
  \item \textit{True or False, for A and for B.} A real transition happens (not just a cut, fade or frozen image). \\ \textit{Answer for each video.}
  \item \textit{True or False, for A and for B.} It starts on the start image. \\ \textit{Answer for each video.}
  \item \textit{True or False, for A and for B.} It leaks content from the intended transition. \\ \textit{A leak: people, objects or background from the intended transition's own scene show up although they are not part of the effect. Things that belong to the effect itself are not a leak.}
\end{enumerate}
The instruction screen adds: the videos play at their own speed and length, then wait for each other and replay together, and a shorter or less smooth video is never a reason to prefer one; the study is anonymous and only the answers are recorded.

\paragraph{VLM judge.} Gemini~2.5~Pro, temperature 0, videos sampled at 8 frames per second, structured JSON output with the evidence field before the answer. Each question is asked in its own call with only the inputs it is defined on. Per pair, two forced-choice calls, (i) transition fidelity and (ii) overall, each from the given frame(s), the reference and the two outputs; each is made twice with the outputs swapped, and a pair counts as a win or a loss only when both orders agree, otherwise as a tie. Per clip, two yes/no calls: (iii) whether a transition effect occurs at all and whether content of the reference's own scenes appears, from the given frame(s), the reference and that clip alone; (iv) whether the clip's first (and last) frame is the given frame, from stills only. A clip without an effect loses both forced-choice questions by rule and is left out of the endpoint count. No system, effect or prompt is ever named to the judge. Call (i) also returns a per-output effect answer, which agrees with call (iii) on 99.8\% of \segue{} clips and 98.6\% of the other systems' clips. Table~\ref{tab:studywtl} gives the counts behind Table~\ref{tab:study}. The four instructions follow verbatim (two-endpoint variant; the one-endpoint variant drops the clauses about the given end frame).

\begin{table}[t]
    \centering
    \caption{\textbf{VLM judge, counts behind Table~\ref{tab:study}.} Wins\,/\,ties\,/\,losses of \segue{} on the two forced-choice questions; a tie is a pair whose choice flipped with the presentation order. Twenty-eight pairs were decided by the no-effect rule.}
    \label{tab:studywtl}
    \footnotesize
    \setlength{\tabcolsep}{6pt}
    \begin{tabular}{lcc}
        \toprule
        \segue{} vs. & Transition fidelity & Overall \\
        \midrule
        \multicolumn{3}{l}{\emph{Transition effect generation: both endpoints given} ($n=38$)} \\
        \refvfx{}~\citep{jones2026tuning}       & 34\,/\,4\,/\,0 & 30\,/\,3\,/\,5 \\
        Base \ltx{}                        & 28\,/\,6\,/\,4 & 24\,/\,7\,/\,7 \\
        \midrule
        \multicolumn{3}{l}{\emph{Visual effect transfer: start endpoint given} ($n=183$)} \\
        \vap{}~\citep{bian2026video}             & 161\,/\,11\,/\,11 & 149\,/\,14\,/\,20 \\
        \vfxmaster{}~\citep{li2025vfxmaster} & 117\,/\,37\,/\,29 & 110\,/\,29\,/\,44 \\
        \refvfx{}~\citep{jones2026tuning}       & 143\,/\,20\,/\,20 & 108\,/\,32\,/\,43 \\
        \bottomrule
    \end{tabular}
\end{table}

\footnotesize
\noindent\textbf{Call (i): transition fidelity.}
\begin{verbatim}
You are an expert judge of video transition effects. You receive
the GIVEN START frame as a still image, the GIVEN END frame as a
still image, a REFERENCE video, and two candidate videos, OUTPUT A
and OUTPUT B. The REFERENCE shows a transition effect between its
own two scenes. Both outputs were asked to reproduce that same
effect on the given frame(s), which show different scenes from the
reference's. No text describes the effect: the REFERENCE is the
only specification.

The outputs may come from different systems. All videos are
time-normalized to a similar duration and sampled at 8 frames per
second. Native frame rates and lengths differ, so held or repeated
frames, choppier motion or a shorter clip must never decide a
question. Do not judge visual quality or how well the first and
last frames are kept; those are judged elsewhere.

First, for each output, state whether a transition effect occurs
at all: "yes" if something beyond a plain cut, crossfade,
dissolve, fade through black or white, blur, blend or a static
image happens; "no" otherwise.

Then answer two questions. Answer each on its own. For each, first
write one sentence of evidence citing timestamps from both
outputs, then choose A or B even when the difference is small,
then state the margin: "none", "slight" or "clear".

1. TRANSITION. Which output better reproduces the REFERENCE's
effect? An output without an effect loses this question. Otherwise
compare the effect itself: the same mechanism and material (smoke,
fire, liquid, particles, a portal, a morph, a shatter...),
entering, spreading and clearing the same way, with similar
timing. The scenes are supposed to differ: different people,
objects or backgrounds are not errors.

2. REFERENCE CONTENT. Which output shows LESS content that belongs
to the REFERENCE's own scenes: its people, objects, backgrounds,
the look of its scenes? The given frame(s) show what the output's
scene should contain; anything from the reference's scenes is
leakage. Material that is part of the effect itself (the smoke,
the fire, the particles) is not leakage. If neither output shows
any, choose either and set the margin to "none".

Return only a JSON object:
{"occurrence": {"A": "yes", "B": "yes"},
 "transition": {"evidence": "...", "choice": "A", "margin":
"clear"},
 "reference_content": {"evidence": "...", "choice": "A", "margin":
"none"}}
\end{verbatim}

\noindent\textbf{Call (ii): overall.}
\begin{verbatim}
You are an expert judge of video transition effects. You receive
the GIVEN START frame as a still image, the GIVEN END frame as a
still image, a REFERENCE video, and two candidate videos, OUTPUT A
and OUTPUT B. The REFERENCE shows a transition effect between its
own two scenes. Both outputs were asked to reproduce that same
effect on the given frame(s), which show different scenes from the
reference's. No text describes the effect: the REFERENCE is the
only specification.

The outputs may come from different systems. All videos are
time-normalized to a similar duration and sampled at 8 frames per
second. Native frame rates and lengths differ, so held or repeated
frames, choppier motion or a shorter clip must never decide the
question.

Question: OVERALL, which output is the better result for this
task, the one you would keep? Weigh what a user of this effect
would care about: that the reference's effect is performed on the
given scene, that the video stays on the given scene and its
frame(s), and that it is free of visual failures a viewer would
notice. An output with no effect at all (a cut, crossfade, fade,
blur or still image) cannot be preferred. Different people,
objects or backgrounds from the reference are not errors: the
scenes are supposed to differ.

First write one sentence of evidence citing timestamps from both
outputs, then choose A or B even when the difference is small,
then state the margin: "none", "slight" or "clear".

Return only a JSON object: {"evidence": "...", "choice": "A",
"margin": "clear"}
\end{verbatim}

\noindent\textbf{Call (iii): effect present and reference leak, per clip.}
\begin{verbatim}
You are an expert judge of video transition effects. You receive
the GIVEN START frame as a still image, the GIVEN END frame as a
still image, a REFERENCE video, and one OUTPUT video. The
REFERENCE shows a transition effect between its own two scenes.
The OUTPUT was asked to reproduce that same effect on the given
frame(s), which show different scenes from the reference's. No
text describes the effect: the REFERENCE is the only
specification.

Videos are sampled at 8 frames per second. Native frame rates and
lengths differ, so held or repeated frames, choppier motion or a
shorter clip never decide an answer. Do not judge how well the
effect is reproduced, how well the given frames are kept, or
visual quality; those are judged elsewhere.

Answer two yes/no questions about the OUTPUT alone. For each,
first write one sentence of evidence citing timestamps (for
example 0:01-0:02), then answer.

1. EFFECT. Does a transition effect occur in the OUTPUT at all?
"yes" if something happens beyond a plain cut, a crossfade or
dissolve, a fade through black or white, a blur or blend between
frames, or a static image. "no" otherwise. The effect does not
need to resemble the reference's to count.

2. LEAK. Does content that belongs to the REFERENCE's own scenes
appear in the OUTPUT: a recognizable person, object, background or
scene of the reference? The given frame(s) show what the OUTPUT's
scene should contain. Material that is part of the effect itself
(smoke, fire, liquid, particles, light) is not leakage. A color
cast or a vague shape is not leakage. The subject copying the pose
or action of the reference's subject is not leakage. If "yes",
name what leaked in the evidence.

Return only a JSON object:
{"effect": {"evidence": "...", "answer": "yes"},
 "leak": {"evidence": "...", "answer": "no"}}
\end{verbatim}

\noindent\textbf{Call (iv): endpoints kept, per clip, stills only.}
\begin{verbatim}
You are an expert judge. You compare still frames only.

You receive the GIVEN START frame and the GIVEN END frame, then
the FIRST frame of the OUTPUT and the LAST frame of the OUTPUT.
The output was supposed to begin on the GIVEN START and end on the
GIVEN END.

Question: does the output's first frame show the GIVEN START, and
does the output's last frame show the GIVEN END? Judge only what
is visible in these stills: the same scene, the same subject and
identity, the same camera framing, the same pose. A frame that
shows all of these still counts as the given frame ("yes") even if
it is softer, blurrier, darker, color-shifted or partly warped, as
long as the scene, subject and pose remain recognizable. Answer
"no" when the frame shows a different scene, a different subject
or identity, a clearly different framing or pose, or is degraded
so far that the given frame is no longer recognizable. Ignore
anything you might guess about the video in between.

For each frame, first write one sentence naming what is the same
and what differs, then answer "yes" or "no".

Return only a JSON object: {"start": {"evidence": "...", "answer":
"yes"}, "end": {"evidence": "...", "answer": "yes"}}
\end{verbatim}
\normalsize
\section{\texorpdfstring{\rev{Ablations}}{Ablations}}
\label{app:ablations}
\paragraph{\rev{Training prompt and attention.}}\label{app:promptattn}
\rev{Table~\ref{tab:ablationpromptattn} separates the two training choices of \segue{}: neutral training prompts and one-way
attention. We train all four combinations on the same data for half the training schedule of \segue{} and evaluate them on a
class-balanced subset of 68 held-out class samples of the evaluation set, with both the neutral (N) and the effect (E) prompt;
the gap $|\Delta|$ between the two measures text dependence. Neutral training prompts have the largest effect: they reduce
$|\Delta|$ from 37.9 to 16.7 with full attention and from 29.3 to 14.5 with one-way attention. One-way attention further reduces
text dependence under both training prompts, from 37.9 to 29.3 with effect prompts and from 16.7 to 14.5 with neutral prompts.
Combined, the \segue{} recipe gives the highest transition fidelity with the neutral prompt (73.9) and the smallest gap (14.5)
of the four ablation models.}

\rev{With full training on the same samples, \segue{} w/o NRG raises transition fidelity to 86.9 and lowers the gap to 3.6, and
NRG improves these further to 91.7 and 1.7. The four ablation models use the training setup of \segue{} (Appendix~\ref{app:train}):
the same training videos, a rank-128 LoRA, AdamW with the learning rate decayed linearly from $10^{-4}$ to $10^{-5}$, and an
effective batch of 8, but 5,000 steps instead of 10,000.}

\begin{table}[H]
    \revfloat
    \centering
    \caption{\textbf{Training prompt and attention ablation} on a class-balanced subset of 68 held-out class samples. N and E: transition fidelity with the neutral and the effect prompt; $|\Delta|$: their gap. Ablation models: half the training steps of the final models.}
    \label{tab:ablationpromptattn}
    \footnotesize
    \setlength{\tabcolsep}{6pt}
    \begin{tabular}{llcccc}
        \toprule
        Attention & Training prompt & Steps & N\,$\uparrow$ & E\,$\uparrow$ & $|\Delta|$\,$\downarrow$ \\
        \midrule
        \tabsec{6}{Ablation models} \\
        Full & Effect & 5k & 52.1 & \textbf{90.0} & 37.9 \\
        Full & Neutral & 5k & 72.6 & 89.3 & 16.7 \\
        One-way & Effect & 5k & 58.9 & 88.2 & 29.3 \\
        One-way & Neutral (\segue{} recipe) & 5k & \textbf{73.9} & 88.4 & \textbf{14.5} \\
        \midrule
        \tabsec{6}{Final models on the same samples} \\
        One-way & Neutral (\segue{} w/o NRG) & 10k & 86.9 & 90.5 & 3.6 \\
        One-way + NRG & Neutral (\segue{}) & 10k & \textbf{91.7} & \textbf{93.4} & \textbf{1.7} \\
        \bottomrule
    \end{tabular}
\end{table}

\paragraph{Guidance weight.}\label{app:guidanceweight}
Table~\ref{tab:weight} sweeps the guidance weight $w$ of \segue{} on each tier. Transition fidelity rises with $w$ on every tier; on the held-out instance and held-out class tiers the seam-free rate gives up a few points at $w = 6$. We use $w = 6$ throughout.

\begin{table}[H]
    \centering
    \caption{\textbf{Guidance weight $w$ by tier.} $w = 1$ is \segue{} w/o NRG, $w = 6$ is \segue{}.}
    \label{tab:weight}
    \footnotesize
    \setlength{\tabcolsep}{4pt}
    \begin{tabular}{llccccc}
        \toprule
        Tier & & \shortstack{Transition\\fidelity\,$\uparrow$} & \shortstack{Motion\\fid.\,$\uparrow$\\($\times$100)} & \shortstack{Action\\KL\,$\downarrow$}
        & \shortstack{Ref\\sim.\,$\uparrow$\\($\times$100)} & \shortstack{Seam-\\free\,\%\,$\uparrow$} \\
        \midrule
        Seen & $w = 1$ & 92.1 & 13.3 & 4.35 & 96.1 & 96.2 \\
             & $w = 1.5$ & 92.8 & 13.9 & 4.09 & 96.2 & \textbf{98.1} \\
             & $w = 3$ & 94.1 & 13.8 & 4.09 & 96.3 & \textbf{98.1} \\
             & $w = 6$ & \textbf{96.1} & \textbf{14.2} & \textbf{3.81} & \textbf{96.3} & \textbf{98.1} \\
        \midrule
        Held-out instance & $w = 1$ & 82.5 & 14.8 & 5.02 & 95.9 & \textbf{95.6} \\
               & $w = 1.5$ & 84.0 & 14.7 & 4.89 & 96.0 & 95.3 \\
               & $w = 3$ & 86.5 & 15.5 & 4.80 & 96.1 & 94.5 \\
               & $w = 6$ & \textbf{87.9} & \textbf{16.3} & \textbf{4.58} & \textbf{96.1} & 92.3 \\
        \midrule
        Held-out class & $w = 1$ & 87.5 & 18.8 & 4.61 & 96.2 & 97.1 \\
                  & $w = 1.5$ & 89.7 & 19.2 & 4.54 & 96.3 & 97.1 \\
                  & $w = 3$ & 92.3 & 20.1 & 4.34 & 96.4 & \textbf{97.3} \\
                  & $w = 6$ & \textbf{93.4} & \textbf{21.0} & \textbf{4.23} & \textbf{96.4} & 94.1 \\
        \bottomrule
    \end{tabular}
\end{table}

\paragraph{\rev{NRG gain by transition type.}}\label{app:nrgtype}
\rev{Table~\ref{tab:nrgtype} splits the gain of NRG by transition type. Procedural transitions such as wipes, warps and
dissolves already score near the top without guidance and gain almost nothing: a cut or a dissolve is what the base model
produces between two scenes by default (Appendix~\ref{app:lerp}), and as a target approaches a dissolve, the guidance term
approaches zero. Creative and EffectData effects start lower and gain the most, up to 9.5 points. NRG thus helps where the
model needs it, without hurting dissolve-like transitions.}

\begin{table}[H]
    \revfloat
    \centering
    \caption{\textbf{NRG gain by transition type}: transition fidelity at $w = 1$ and $w = 6$ (neutral prompt) and the gain, paired per sample (mean $\pm$ s.e.). Procedural: 10 held-out shader families, outside the evaluation set.}
    \label{tab:nrgtype}
    \footnotesize
    \setlength{\tabcolsep}{4pt}
    \begin{tabular}{lcccc}
        \toprule
        Transition type & Samples & $w = 1$ & $w = 6$ & NRG gain \\
        \midrule
        \tabsec{5}{Creative, all tiers} \\
        Smoke / gas / fire / explosion & 54 & 90.6 & 95.5 & $+4.9 \pm 1.3$ \\
        Particles and elements (rain, petals, water, earth, air) & 48 & 84.0 & 90.5 & $+6.5 \pm 1.4$ \\
        Geometry / wireframe / point-cloud restyle & 38 & 76.8 & 86.2 & $+9.5 \pm 2.4$ \\
        Camera / motion / portal & 62 & 85.6 & 93.8 & $+8.2 \pm 1.8$ \\
        Appearance and character restyle & 80 & 83.2 & 85.0 & $+1.8 \pm 1.3$ \\
        \textbf{All creative} & 282 & \textbf{84.4} & \textbf{90.0} & $\mathbf{+5.6 \pm 0.7}$ \\
        \midrule
        \tabsec{5}{EffectData, held-out class tier} \\
        \textbf{All EffectData effects} & 102 & \textbf{90.4} & \textbf{95.9} & $\mathbf{+5.4 \pm 0.8}$ \\
        \midrule
        \tabsec{5}{Procedural, held-out shader families} \\
        Wipe / warp / zoom & 50 & 97.1 & 96.6 & $-0.5 \pm 0.4$ \\
        Blur dissolve / grayscale fade & 20 & 98.8 & 98.7 & $+0.0 \pm 0.4$ \\
        Glitch / mosaic / burn & 30 & 98.2 & 99.1 & $+0.9 \pm 0.6$ \\
        \textbf{All procedural} & 100 & \textbf{97.8} & \textbf{97.8} & $\mathbf{+0.0 \pm 0.3}$ \\
        \bottomrule
    \end{tabular}
\end{table}

\paragraph{\rev{Negative reference.}}\label{app:nulldesigns}
\rev{Table~\ref{tab:nulldesigns} compares the dissolve with other negatives: an empty reference
(Figure~\ref{fig:nullcmp}), the reference with its frames shuffled, and the reference of another effect. Every negative
raises transition fidelity over \segue{} without guidance by a similar amount. Transition fidelity does not see seams or
copying of the reference's scenes, however: among the negatives, the dissolve keeps the hand-off cleanest (the highest
seam-free rate) and copies the reference's scenes the least (the lowest reference-endpoint similarity).}

\begin{table}[H]
    \revfloat
    \centering
    \caption{\textbf{Negative references for NRG}, on the held-out class samples of Table~\ref{tab:ablationpromptattn}. Ref.-endpoint sim.: similarity of the generated transition to the reference's own endpoint frames; similarity to the reference's effect is wanted, similarity to its endpoints is copying.}
    \label{tab:nulldesigns}
    \footnotesize
    \setlength{\tabcolsep}{4pt}
    \begin{tabular}{lccccc}
        \toprule
        Negative reference & \shortstack{Transition\\fidelity\,$\uparrow$} & \shortstack{Motion\\fid.\,$\uparrow$\\($\times$100)} & \shortstack{Ref\\sim.\,$\uparrow$\\($\times$100)} & \shortstack{Seam-\\free\,\%\,$\uparrow$} & \shortstack{Ref.-endpoint\\sim.\,$\downarrow$\\($\times$100)} \\
        \midrule
        None (\segue{} w/o NRG, $w = 1$) & 86.9 & 19.0 & 96.1 & 98.5 & 24.4 \\
        \midrule
        Dissolve (ours) & 91.7 & 22.0 & 96.4 & \textbf{95.6} & \textbf{29.1} \\
        Empty (reference dropped, Figure~\ref{fig:nullcmp}) & 91.6 & \textbf{22.7} & \textbf{96.5} & 91.2 & 31.2 \\
        Shuffled reference frames & \textbf{92.1} & 21.2 & 96.4 & 91.2 & 29.4 \\
        Another effect's reference & 90.5 & 20.9 & 96.4 & 88.2 & 29.7 \\
        \bottomrule
    \end{tabular}
\end{table}

\rev{Table~\ref{tab:ablationtiers} gives the ablations on all three tiers.}

\paragraph{\rev{Plain LoRA.}}\label{app:plainlora}
\rev{Plain LoRA was our first experiment, where we tested and verified text dependence. It fine-tunes the same \ltx{}
backbone with LoRA (rank 32, learning rate $2\times10^{-4}$, 5,000 steps) on 139 creative-transition clips covering 26 effects.
Like \segue{}, it is trained with neutral prompts that describe only the scenes. Unlike \segue{}, it uses full attention
between the reference and the target, with no null-reference guidance. With neutral-prompt training, Plain LoRA becomes
less dependent on text, with a gap of only 7.6 in transition fidelity on seen effects, compared with 20.7 to 35.6 for the
other baselines (Table~\ref{tab:main}). However, on held-out effect classes, its transition fidelity drops to 61.9 and the
gap grows to 19.6, showing poor generalization and greater reliance on text. This motivated our cross-paired dataset and
the other components of \segue{}, which achieves a transition fidelity of 93.4 with a gap of just 1.2 on the same classes.}

\begin{table}[H]
    \revfloat   %
    \centering
    \caption{\textbf{Ablations on the three tiers}, both tasks included. N, E, and $|\Delta|$ as in Table~\ref{tab:main}. The held-out class tier pools the two tasks, which Table~\ref{tab:main} reports separately.}
    \label{tab:ablationtiers}
    \scriptsize
    \setlength{\tabcolsep}{2pt}
    \begin{tabular*}{\linewidth}{@{\extracolsep{\fill}}lccc!{\hspace{3pt}}ccc!{\hspace{3pt}}ccc!{\hspace{3pt}}ccc!{\hspace{3pt}}ccc@{}}
        \toprule
        & \multicolumn{3}{c}{Transition fidelity\,$\uparrow$} & \multicolumn{3}{c}{Motion fid.\,($\times$100)\,$\uparrow$} & \multicolumn{3}{c}{Action KL\,$\downarrow$} & \multicolumn{3}{c}{Ref sim.\,($\times$100)\,$\uparrow$} & \multicolumn{3}{c}{Seam-free\,\%\,$\uparrow$} \\
        \cmidrule(lr){2-4}\cmidrule(lr){5-7}\cmidrule(lr){8-10}\cmidrule(lr){11-13}\cmidrule(lr){14-16}
         & N & E & $|\Delta|$ & N & E & $|\Delta|$ & N & E & $|\Delta|$ & N & E & $|\Delta|$ & N & E & $|\Delta|$ \\
        \midrule
        \tabsec{16}{Seen} \\
        Plain LoRA & 85.2 & 92.8 & 7.6 & \textbf{17.8} & \textbf{15.4} & 2.4 & 4.25 & 4.23 & \textbf{0.02} & 95.9 & 96.0 & 0.1 & 94.2 & 94.2 & \textbf{0.0} \\[0.4ex]
        \midrule
        \segue{} w/o NRG & 92.1 & 95.5 & 3.4 & 13.3 & 13.6 & \textbf{0.3} & 4.35 & 4.31 & 0.04 & 96.1 & 96.1 & \textbf{0.0} & 96.2 & 96.2 & \textbf{0.0} \\
        \addlinespace[2pt]
        \segue{} & \textbf{96.1} & \textbf{96.7} & \textbf{0.6} & 14.2 & 14.5 & \textbf{0.3} & \textbf{3.81} & \textbf{3.84} & 0.03 & \textbf{96.3} & \textbf{96.4} & 0.1 & \textbf{98.1} & \textbf{100.0} & 1.9 \\
        \midrule
        \tabsec{16}{Held-out instance} \\
        Plain LoRA & 72.6 & 84.3 & 11.7 & 15.3 & 14.4 & 0.9 & 5.52 & 5.06 & 0.46 & 95.5 & 95.9 & 0.4 & \textbf{97.8} & \textbf{97.4} & \textbf{0.4} \\[0.4ex]
        \midrule
        \segue{} w/o NRG & 82.5 & 87.6 & 5.1 & 14.8 & 14.7 & \textbf{0.1} & 5.02 & 4.74 & 0.28 & 95.9 & 96.1 & 0.2 & 95.6 & 94.2 & 1.4 \\
        \addlinespace[2pt]
        \segue{} & \textbf{87.9} & \textbf{89.9} & \textbf{2.0} & \textbf{16.3} & \textbf{16.6} & 0.3 & \textbf{4.58} & \textbf{4.47} & \textbf{0.11} & \textbf{96.1} & \textbf{96.2} & \textbf{0.1} & 92.3 & 91.2 & 1.1 \\
        \midrule
        \tabsec{16}{Held-out class} \\
        Plain LoRA & 61.9 & 81.5 & 19.6 & 15.4 & 15.8 & 0.4 & 5.35 & 4.50 & 0.85 & 95.3 & 96.0 & 0.7 & 95.9 & \textbf{97.5} & 1.6 \\[0.4ex]
        \midrule
        \segue{} w/o NRG & 87.5 & 91.9 & 4.4 & 18.8 & 19.0 & 0.2 & 4.61 & 4.44 & 0.17 & 96.2 & 96.4 & 0.2 & \textbf{97.1} & 96.8 & 0.3 \\
        \addlinespace[2pt]
        \segue{} & \textbf{93.4} & \textbf{94.6} & \textbf{1.2} & \textbf{21.0} & \textbf{21.0} & \textbf{0.0} & \textbf{4.23} & \textbf{4.13} & \textbf{0.10} & \textbf{96.4} & \textbf{96.5} & \textbf{0.1} & 94.1 & 93.9 & \textbf{0.2} \\
        \bottomrule
    \end{tabular*}
\end{table}

\section{Additional results}
\label{app:perop}

\begin{figure}[tb]
  \centering
  \includegraphics[width=\linewidth]{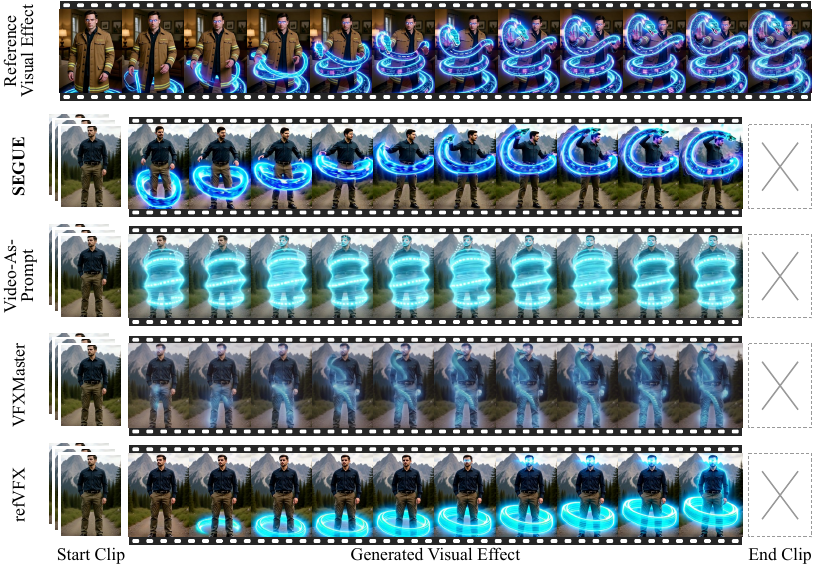}
  \caption{\textbf{Qualitative comparison}, start endpoint given. Top: the reference; below: one system per row, frames in time order.}
  \label{fig:qualitativevet}
\end{figure}

Figure~\ref{fig:qualitativevet} compares the systems of Table~\ref{tab:main} with the start endpoint given, in the layout of Figure~\ref{fig:qualitative}.

\begin{figure}[p]
  \centering
  \includegraphics[width=\linewidth]{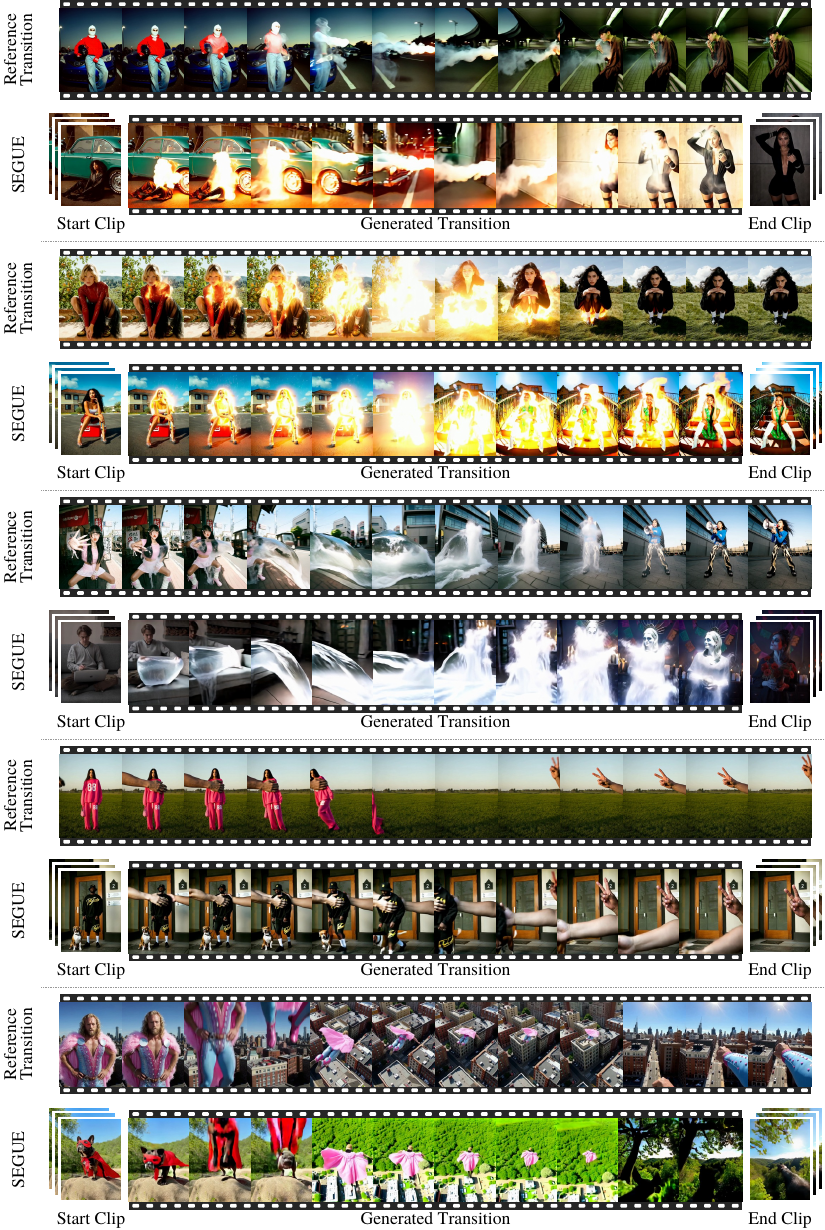}
  \caption{\textbf{More results of \segue{}}, both endpoints given.}
  \label{fig:galteg}
\end{figure}

\begin{figure}[p]
  \centering
  \includegraphics[width=0.885\linewidth]{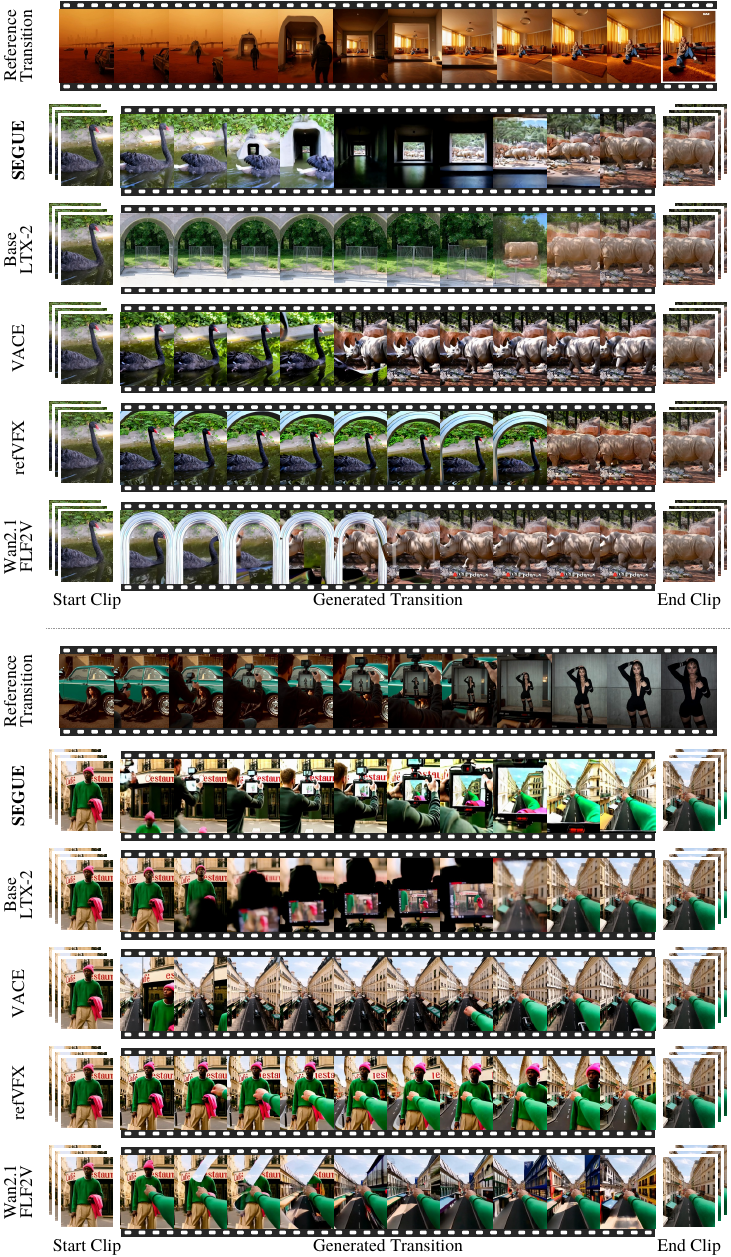}   %
  \caption{\textbf{More comparisons}, held-out class tier, both endpoints given, on samples where the VLM judge preferred \segue{}.}
  \label{fig:cmpteg}
\end{figure}

\begin{figure}[p]
  \centering
  \includegraphics[width=\linewidth]{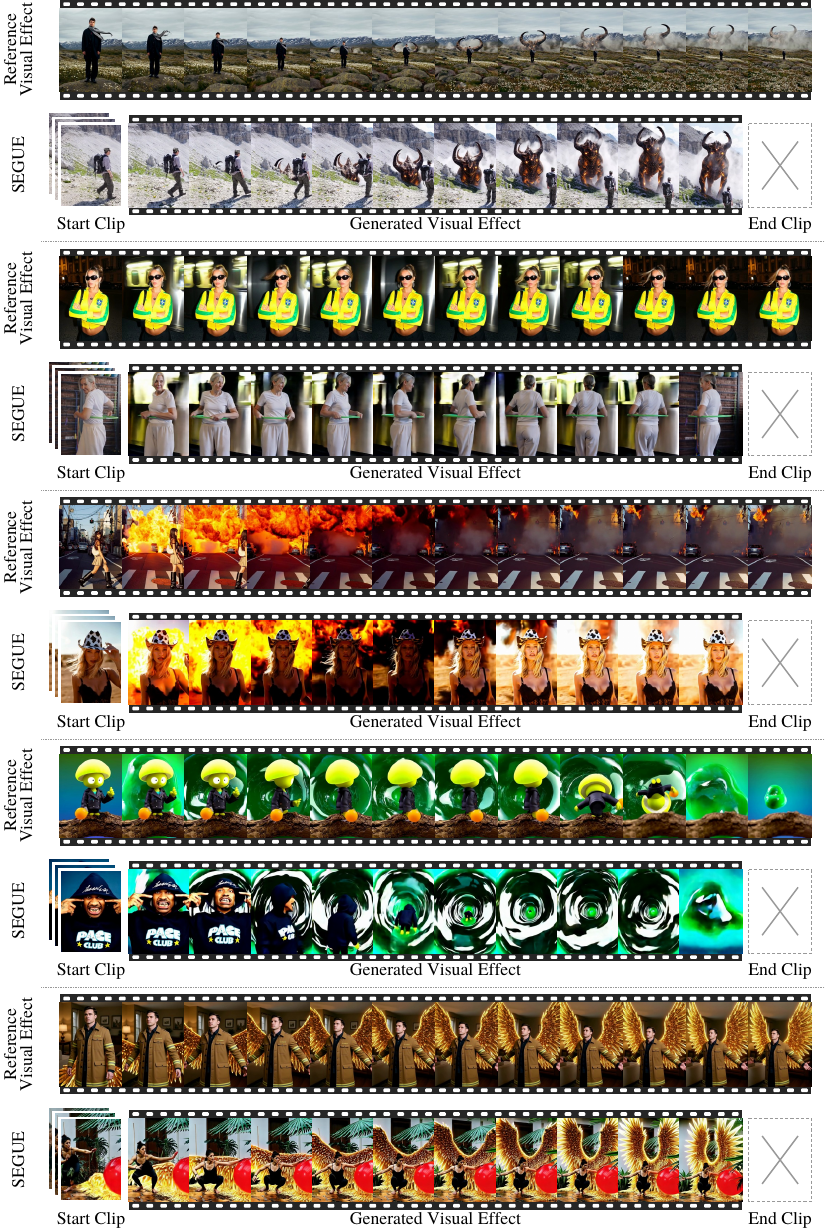}
  \caption{\textbf{More results of \segue{}}, start endpoint given.}
  \label{fig:galvet}
\end{figure}

\begin{figure}[p]
  \centering
  \includegraphics[width=\linewidth]{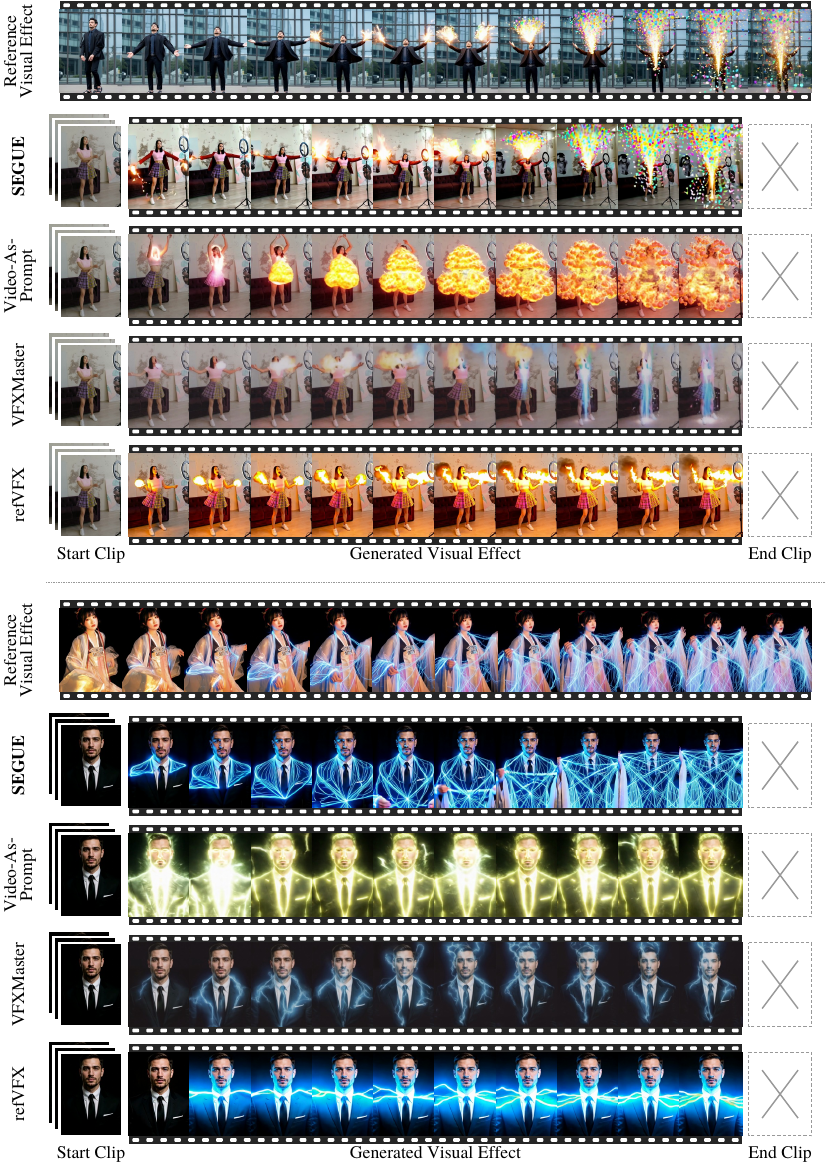}
  \caption{\textbf{More comparisons}, held-out class tier, start endpoint given, on samples where the VLM judge preferred \segue{}.}
  \label{fig:cmpvet}
\end{figure}

Figures~\ref{fig:galteg} to~\ref{fig:cmpvet} show more results of \segue{} and more comparisons, in the layouts of Figures~\ref{fig:hero} and~\ref{fig:qualitative}.

\ifarxiv
\textbf{Videos of these and more results are on the project page}, \url{\projectpage}; refer to them for a better grasp of the results of \segue{} and prior art.
\else
\textbf{The supplementary material shows videos with qualitative examples.}
Refer to those for a better grasp of the results of \segue{} and prior art.
Open the static HTML file (\url{index.html}) using a modern web browser that supports HTML5 video rendering.
While Google Chrome and Mozilla Firefox are recommended, please note that native video decoding often depends on OS-specific media frameworks and codecs. If playback issues arise, verify your browser’s core video decoding capabilities.
For advanced users troubleshooting performance or rendering blocks, consider disabling browser extensions related to video management, ad-blocking, or memory optimization (such as Chrome Memory Saver or Firefox Efficiency Mode).
\fi

\begin{figure}[p]
  \centering
  \includegraphics[width=\linewidth]{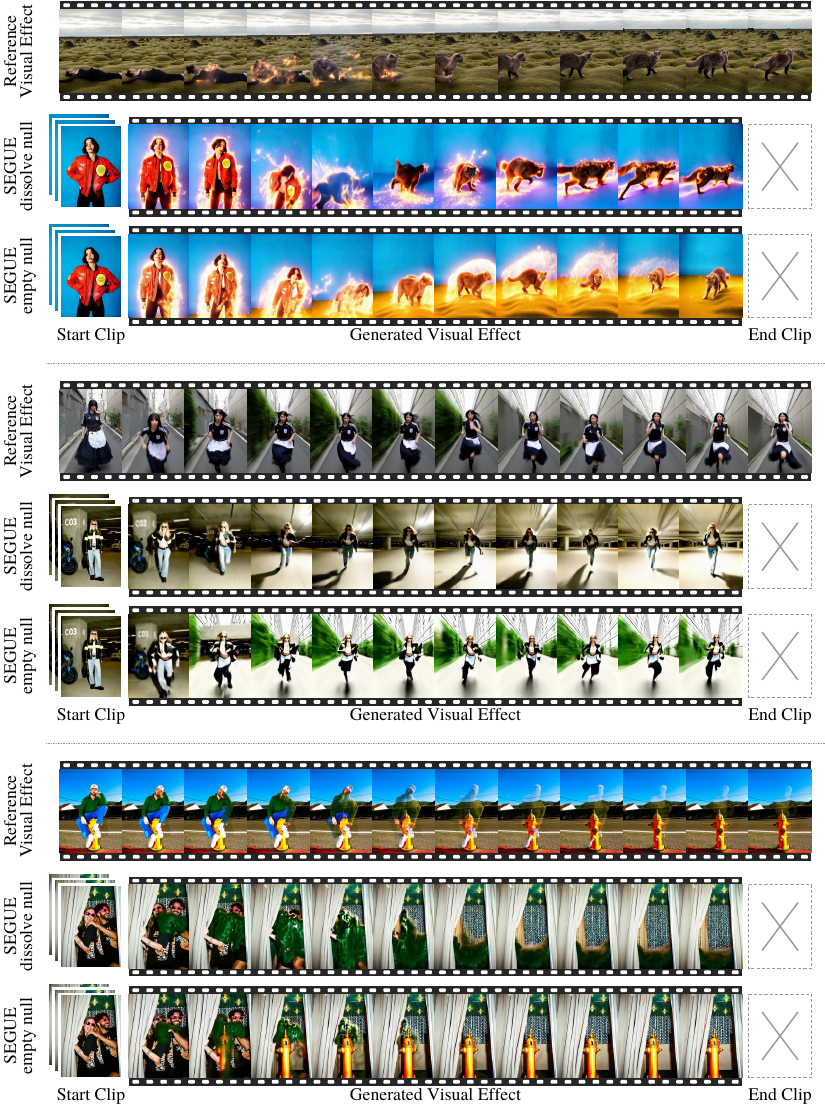}
  \caption{\textbf{Negative reference: empty null vs.\ dissolve null.} Each block shows the reference clip (top strip) and \segue{} run twice with the \emph{same} reference, endpoints, neutral prompt, seed~42 and guidance $w=6$. Only the null reference changes. With the \emph{empty} null (the reference dropped from the null pass) the guidance amplifies the reference's own scenes along with its effect, so reference content leaks into the output: the ground (top block), the running scene (middle) and the fire hydrant (bottom). With \segue{}'s \emph{dissolve} null (a dissolve of the reference's own endpoints) only the effect is amplified and the start clip's scene is kept.}
  \label{fig:nullcmp}
\end{figure}

\paragraph{Negative reference.}
Section~\ref{sec:inference} contrasts \segue{}'s dissolve null with an empty null that simply drops the reference from the null pass. Figure~\ref{fig:nullcmp} makes the difference concrete on three examples: with the empty null the guidance amplifies the reference's own scenes along with its effect, so the ground, the running scene, and the fire hydrant of the reference leak into the output. With the dissolve null the start clip's scene is kept and only the effect is amplified.

\begin{figure}[t]
  \centering
  \includegraphics[width=\linewidth]{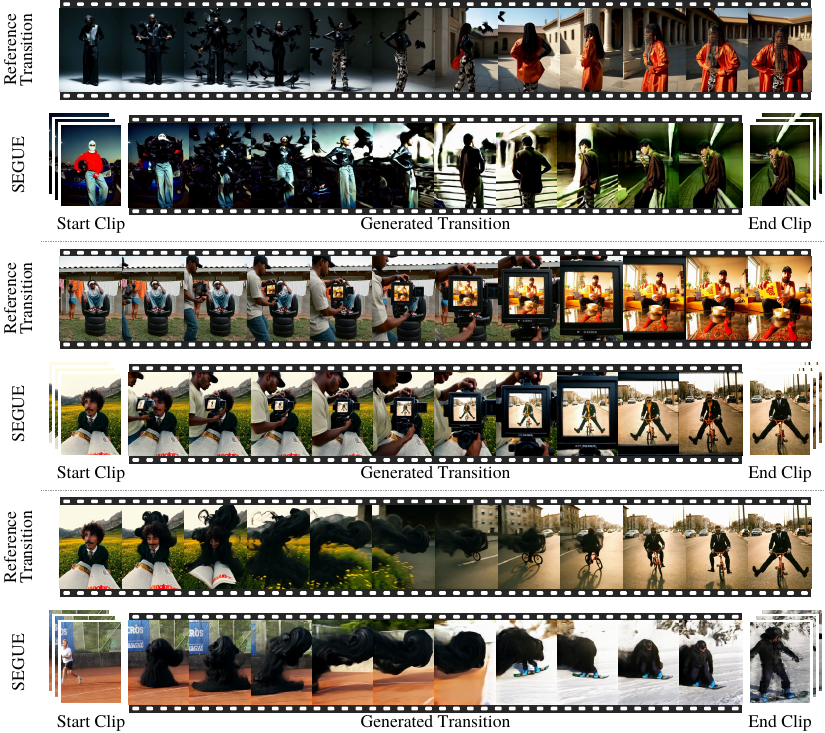}
  \caption{\textbf{Failure cases of \segue{}}, both endpoints given. Top two rows: scenes of the reference leak into the output. Bottom row: the appearance of the endpoints is not preserved through the transition.}
  \label{fig:limitations}
\end{figure}

\paragraph{Failure cases.}
Figure~\ref{fig:limitations} shows the two failure modes of Section~\ref{sec:conclusion} on three samples: in the first two, scenes of the reference leak into the output; in the third, whose endpoint shows intense motion, the endpoints are not preserved.

\section{Evaluation dataset}
\label{app:grid}
\begin{table}[H]
    \centering
    \caption{\textbf{Evaluation dataset by tier.} Endpoints are shared across tiers, so the total counts distinct endpoints.}
    \label{tab:grid}
    \footnotesize
    \setlength{\tabcolsep}{6pt}
    \begin{tabular}{lrrrr}
        \toprule
        & & & \multicolumn{2}{c}{Samples} \\
        \cmidrule(lr){4-5}
        Tier & Effect types & Distinct endpoints & Both endpoints & Start endpoint \\
        \midrule
        Seen              & 13 & 26  & 4  & 22  \\
        Held-out instance & 25 & 82  & 32 & 105 \\
        Held-out class    & 51 & 150 & 38 & 183 \\
        \midrule
        Total             & 76 & 204 & 74 & 310 \\
        \bottomrule
    \end{tabular}
\end{table}

\paragraph{Tiers.}
The evaluation dataset has 384 samples in three tiers, defined by how the reference clip relates to the training set of \segue{}. In the \emph{seen} tier the reference clip itself was trained on. In the \emph{held-out instance} tier the reference clip was not, but other clips of its effect were. In the \emph{held-out class} tier the whole effect was held out: none of its 51 effects, 68 reference clips, or 150 endpoints appears in the training set. The held-out class tier is the evaluation dataset of Section~\ref{sec:evaldata}; the other two tiers serve the appendix tables. Table~\ref{tab:grid} lists the three tiers.

\paragraph{Samples.}
A sample is a reference clip together with the target endpoints: both endpoints for two-sided \ctt{}, or the start alone for single-sided visual effect transfer. The endpoints are of three kinds, so that the same reference is tested on scenes at three distances from it: a scene from another clip of the reference's own effect, never the reference clip itself (124 samples); a scene from a clip of another effect of our sources (130 samples); and a clip from DAVIS or HumanVid that no effect was ever applied to (130 samples). The seen tier uses only the first kind. Endpoints are shared across tiers, so the 204 endpoints of Table~\ref{tab:grid} are distinct.

\paragraph{Held-out class tier.}
Its 51 effects are the 17 creative transitions held out from training, each with two reference clips and six to eight samples, and the 34 EffectData effects, each with one reference clip and three samples, one of each endpoint kind. The EffectData effects are chosen as 17 couples that share a start subject, so the same subject appears under two different effects, and their endpoints are 51 EffectData subjects and 17 HumanVid clips. EffectData samples are 81 frames long, the native length of its clips; all other samples are 121 frames.

\paragraph{Generation.}
Every sample is run with two seeds. Each of our variants generates all 384 samples with the neutral prompt, 768 videos, and once more with the effect prompt for the E and $|\Delta|$ columns of Tables~\ref{tab:main} and~\ref{tab:ablationtiers}. Previous approaches generate the 183 start-only samples of the held-out class tier ($n = 366$ per system), and the adapted baselines its 38 two-endpoint samples ($n = 76$ per system).
\section{Distance from linear interpolation}
\label{app:lerp}
Section~\ref{sec:inference} states that the base model, given two endpoints and a prompt that describes only the two scenes, mostly produces a cut or a dissolve. In our qualitative inspection this holds almost always: given both endpoints, the base model falls back to a traditional transition such as a dissolve, a fade, a blur or a cut. In pixel space both belong to the linear interpolation family $\lerp$,
\begin{equation}
    V_\tau = \big(1 - \alpha(\tau)\big)\, a + \alpha(\tau)\, b,
    \qquad \text{for a monotonic schedule } \alpha(\tau) \in [0, 1],
    \label{eq:lerp}
\end{equation}
where $\tau \in [0, 1]$ is normalized time, $a$ is the last frame of the start, and $b$ the first frame of the end. A member of $\lerp$ only blends the two given frames: it keeps the scenes of both endpoints and adds no motion and no material of its own; the dissolve of Eq.~\eqref{eq:null} is the member with $\alpha(\tau) = \tau$. Table~\ref{tab:lerp} measures how far videos sit from $\lerp$: the outputs of the base \ltx{} sit less than half as far from $\lerp$ as real transitions do.
\begin{table}[H]   %
    \centering
    \caption{\textbf{Distance from linear interpolation.} Mean $\pm$ s.d. over $n$ clips.}
    \label{tab:lerp}
    \footnotesize
    \setlength{\tabcolsep}{3pt}
    \begin{tabular}{lcc}
        \toprule
                                & $n$ & Distance from $\lerp$ \\
        \midrule
        Real transitions        & 73 & $0.54 \pm 0.15$ \\
        Base \ltx{} transitions & 90 & $0.24 \pm 0.11$ \\
        \bottomrule
    \end{tabular}
\end{table}

\end{document}